\def\buildpreprint{}
\documentclass{article}

\newif\ifpreprint
\ifdefined\buildpreprint \preprinttrue \fi
\ifpreprint
\usepackage{arxiv,times}
\else
\usepackage{iclr2027_conference,times}
\fi

\usepackage{amsmath,amsfonts,bm}

\def\eqref#1{equation~\ref{#1}}

\def\1{\bm{1}}

\def\vzero{{\bm{0}}}

\def\va{{\bm{a}}}

\def\vv{{\bm{v}}}

\def\vz{{\bm{z}}}

\def\mI{{\bm{I}}}

\DeclareMathAlphabet{\mathsfit}{\encodingdefault}{\sfdefault}{m}{sl}
\SetMathAlphabet{\mathsfit}{bold}{\encodingdefault}{\sfdefault}{bx}{n}

\usepackage[utf8]{inputenc} %
\usepackage[T1]{fontenc}    %
\usepackage[table]{xcolor}  %
\usepackage[colorlinks=true,allcolors=blue!60!black]{hyperref} %
\usepackage{url}            %
\usepackage{booktabs}       %
\usepackage{microtype}      %
\usepackage{xspace}         %
\usepackage{subcaption}     %
\usepackage{graphicx}
\usepackage{float}
\usepackage{pifont}
\usepackage{mathtools,amsthm,amssymb,multirow}
\usepackage{enumitem}
\usepackage{tabularx}

\usepackage[capitalize,nameinlink]{cleveref}
\crefname{section}{Sec.}{Secs.}
\crefname{appendix}{App.}{Apps.}
\crefname{figure}{Fig.}{Figs.}

\usepackage{tikz,pgfplots}
\usetikzlibrary{decorations.pathreplacing,arrows.meta,positioning,calc,shapes.arrows,fit,backgrounds}
\graphicspath{{./}{./figures/}{./attention_diagrams/}}

\newlength{\figheight}

\newcommand{\method}{LeAVJEPA\xspace}

\newcommand{\cmark}{\textcolor{green!60!black}{\ding{51}}}
\newcommand{\xmark}{\textcolor{red!80!black}{\ding{55}}}

\makeatletter
\renewcommand\paragraph{\@startsection{paragraph}{4}{\z@}%
  {.25ex \@plus .1ex \@minus .1ex}%
  {-1em}%
  {\normalfont\normalsize\bfseries}}
\makeatother

\title{\method: A Minimalist Architecture for \\
Audio-Visual Self-Supervised Learning}

\ifpreprint
\author{Benjamin Robson\quad Santeri Mentu\quad Wenshuai Zhao\quad Arno Solin \\
ELLIS Institute Finland \hspace{3em} Aalto University
}
\else
\author{Benjamin Robson, Santeri Mentu, Wenshuai Zhao \& Arno Solin \\
ELLIS Institute Finland and Department of Computer Science \\
Aalto University \\
Espoo, Finland \\
\texttt{benjamin.robson@aalto.fi}
}
\fi
\ifpreprint \projectpage{https://aaltoml.github.io/LeAVJEPA/} \fi

\begin{document}
\maketitle
\begin{abstract}
Prior audio-visual self-supervised learning methods rely on mechanisms
such as EMA target encoders, prediction heads, reconstruction decoders, and
contrastive losses. We introduce \method, the first audio-visual encoder trained
under LeJEPA's collapse-free objective. A single early-fusion Vision
Transformer processes audio, video, and joint audio-video inputs. Modality
dropout treats a missing modality as another view of the same event, making
cross-modal alignment implicit in the objective. The model aligns global
embeddings with modality-specific local embeddings, and SIGReg prevents
representational collapse. A controlled ablation identifies modality dropout
as the key mechanism for audio-visual alignment. Despite the architectural
simplicity, \method reaches 36.0 mAP on AudioSet-20K and 91.3\%
accuracy on ESC-50 under frozen evaluation. After fine-tuning, it reaches 61.1\% accuracy on VGGSound,
and its embeddings support zero-shot audio-visual retrieval.
\end{abstract}

\addtocontents{toc}{\protect\setcounter{tocdepth}{0}}

\begin{figure}[b!]
  \centering\scriptsize
  \setlength{\figheight}{37mm}
  \providecommand{\avassetpath}{figures/method_diagram/assets/}
  \vspace*{-1em}
  \resizebox{\linewidth}{!}{%
    \input{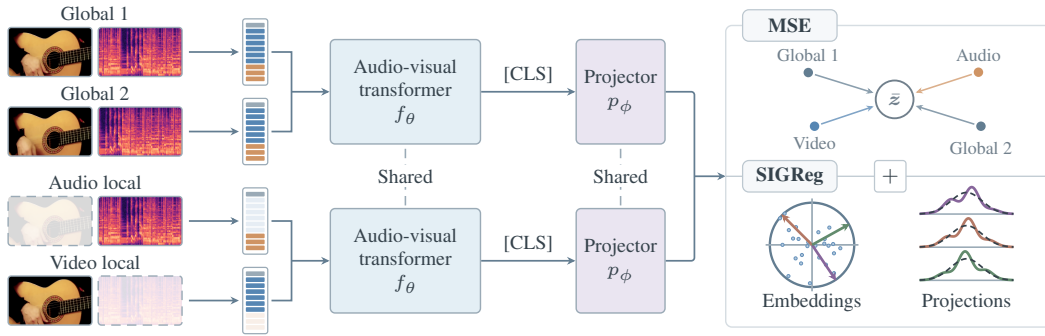}%
  }
  \caption{\textbf{\method connects sound and vision by dropping a modality.}
    One shared encoder aligns audio-only and video-only views with joint audio-video views.
    MSE brings embeddings together; SIGReg prevents collapse, without a teacher, predictor, or contrastive negatives.}
  \label{fig:teaser}
  \label{fig:pipeline}
\end{figure}

\section{Introduction}
\label{sec:intro}
Self-supervised learning (SSL) has become the dominant paradigm for learning representations from unlabelled data in vision~\citep{he2022mae,oquab2024dinov2}, audio~\citep{huang2022audiomae}, and video~\citep{bardes2024vjepa}. Video is a natural source of paired multimodal data: every clip pairs a visual stream with an audio stream that records the same event, and their co-occurrence provides supervision at no annotation cost~\citep{arandjelovic2017l3,korbar2018avts}. The pairing is also imperfect, since sound sources move off-screen and visible objects are often silent. A good audio-visual representation should therefore draw on both modalities, yet remain useful when either one is missing or uninformative.

Existing audio-visual SSL methods combine audio and video through several different training objectives. AV-MAE~\citep{georgescu2023avmae}, CAV-MAE~\citep{gong2023cavmae}, MAViL~\citep{huang2023mavil}, and CAV-MAE Sync~\citep{araujo2025cavmaesync} reconstruct masked spectrogram and pixel patches through decoders, most add InfoNCE terms over cross-modal negatives, and MAViL adds a second self-training stage. Alternatively, predicting in latent space focuses learning on representations and avoids reconstructing low-level input detail~\citep{assran2023ijepa,bardes2024vjepa}. MJEPA~\citep{teotia2026mjepa} takes this route with a shared encoder, but retains an EMA target encoder and a predictor network.
This raises the question of \emph{how many of these components are necessary}. LeJEPA~\citep{balestriero2025lejepa} showed for single-modality vision that a symmetric view-invariance loss plus SIGReg, a regularizer that drives embeddings toward an isotropic Gaussian, trains stably with no teacher, predictor, stop-gradient, or negatives. However, whether the same holds for audio and video is unclear. When every view contains both modalities, the encoder can minimise the invariance loss using only the dominant modality and largely ignore the other. Learning a cross-modal representation therefore depends on views that separate the modalities, such that audio-only and video-only views of the same clip are mapped to a common embedding.

We introduce \textbf{\method}, a minimalist audio-visual JEPA that trains a single shared early-fusion Vision Transformer with only an invariance loss and SIGReg, treating paired audio and video as partial observations of the same event. The encoder maps each audio, video, or joint input to a sequence of tokens (\cref{fig:pipeline}), where global views contain both modalities and \emph{modality dropout} makes local views audio-only or video-only. The invariance loss pulls each local-view embedding toward the centre of the global views, requiring the encoder to infer the joint representation from either modality alone. The views determine which modalities the encoder observes, while the common centre defines the representation to be predicted. Cross-modal prediction thus emerges implicitly in latent space, without a teacher, predictor, decoder, or negatives.

On VGGSound and AudioSet, \method reaches \textbf{61.1\%} accuracy after fine-tuning and \textbf{36.0 mAP} on AudioSet-20K with a frozen attentive probe, and its embeddings support audio$\leftrightarrow$video retrieval without retrieval-specific training. These results show that the shared-encoder approach is effective, although the strongest published methods still perform better. A controlled ablation identifies modality-specific views as central to representation quality. Masked local views nearly halve linear-probe accuracy, and joint local views reduce it to the level of pretraining on audio alone.

Together, these findings support three contributions:
\begin{itemize}[leftmargin=1.5em]
  \item \textbf{A minimalist audio-visual JEPA} that trains a single shared early-fusion encoder with a symmetric invariance loss and SIGReg, with no reconstruction decoder, EMA teacher, predictor, contrastive negatives, or multi-stage schedule (\cref{tab:comparison}).
  \item \textbf{Modality dropout for partial-view learning}, which makes cross-modal prediction an implicit property of the invariance objective rather than a separate loss or network (\cref{sec:cross_modal}).
  \item \textbf{An empirical study} that first isolates the contribution of view construction through a controlled ablation, showing that modality-specific local views substantially outperform masked, joint, audio-only, and no-local-view alternatives (\cref{sec:view_ablation}), and then evaluates frozen transfer, zero-shot retrieval, and end-to-end fine-tuning with ViT-B and ViT-L backbones on VGGSound and AudioSet-2M (\cref{sec:experiments}).
\end{itemize}

\section{Related Work}
\label{sec:related}
\paragraph{Audio-visual self-supervised learning} Early work used audio-visual correspondence~\citep{arandjelovic2017l3} and temporal synchrony~\citep{korbar2018avts} as pretext tasks; AVID-CMA~\citep{morgado2021avid}, XDC~\citep{alwassel2020xdc}, and VATT~\citep{akbari2021vatt} then aligned the two modalities with InfoNCE~\citep{oord2018cpc} or clustering objectives. A second line masks and reconstructs raw inputs: AV-MAE~\citep{georgescu2023avmae}, CAV-MAE~\citep{gong2023cavmae}, MAViL~\citep{huang2023mavil}, and CAV-MAE Sync~\citep{araujo2025cavmaesync} extend MAE~\citep{he2022mae} to paired audio and video, in most cases with an added contrastive term. These approaches establish reconstruction and explicit alignment as effective learning signals. We study how far a common-centre objective can go when modality dropout supplies the partial observations.

\paragraph{Audio-only and video-only SSL backbones} Single-modality methods establish how to turn sound and video into useful representations. wav2vec\,2.0~\citep{baevski2020wav2vec2} and HuBERT~\citep{hsu2021hubert} learn from waveforms; AST~\citep{gong2021ast}, SSAST~\citep{gong2022ssast}, AudioMAE~\citep{huang2022audiomae}, and BEATs~\citep{chen2022beats} operate on spectrograms. ViViT~\citep{arnab2021vivit} and TimeSformer~\citep{bertasius2021timesformer} develop transformer architectures for video. Our architecture uses modality-specific tokenization to bring both streams into one transformer.

\paragraph{Joint-embedding SSL without reconstruction} BYOL~\citep{grill2020byol} and SimSiam~\citep{chen2021simsiam} match embeddings across views without negatives, and DINO~\citep{caron2021dino} and DINOv2~\citep{oquab2024dinov2} add multi-crop self-distillation. A complementary line replaces the asymmetric teacher with an explicit distribution-level regularizer: VICReg~\citep{bardes2022vicreg} and Barlow Twins~\citep{zbontar2021barlowtwins} constrain the embedding covariance and W-MSE~\citep{ermolov2021wmse} whitens it. However, these methods constrain only the second-order statistics of the embedding distribution. \method instead inherits SIGReg~\citep{balestriero2025lejepa}, which matches the full distribution to an isotropic Gaussian using a sliced characteristic-function test.

\begin{table}[t]
  \centering
  \setlength{\abovecaptionskip}{0pt}
  \setlength{\belowcaptionskip}{0pt}
  \caption{Comparison of audio-visual SSL methods. \method is the only method that requires none of the listed components.}
  \label{tab:comparison}  
  \begingroup
  \scriptsize
  \setlength{\tabcolsep}{3pt}
  \renewcommand{\arraystretch}{1.2}
  \resizebox{\linewidth}{!}{%
  \begin{tabular}{lccccccl}
    \toprule
    \textbf{Method}
      & \shortstack{\textbf{Single shared}\\\textbf{encoder}}
      & \shortstack{\textbf{No raw-input}\\\textbf{recon.}}
      & \shortstack{\textbf{No contr.}\\\textbf{negatives}}
      & \shortstack{\textbf{No} \\ \textbf{decoder}}
      & \shortstack{\textbf{No teacher}\\\textbf{/ EMA}}
      & \shortstack{\textbf{No} \\ \textbf{predictor}}
      & \shortstack{\textbf{Training}\\\textbf{objective}} \\
    \midrule
    AV-MAE~\citep{georgescu2023avmae}
      & \xmark & \xmark & \cmark & \xmark & \cmark & \cmark & Recon. \\
    CAV-MAE~\citep{gong2023cavmae}
      & \xmark & \xmark & \xmark & \xmark & \cmark & \cmark & Recon.\,+\,contr. \\
    MAViL~\citep{huang2023mavil}
      & \xmark & \xmark & \xmark & \xmark & \xmark & \cmark & Raw/latent recon.\,+\,contr. \\
    CAV-MAE Sync~\citep{araujo2025cavmaesync}
      & \xmark & \xmark & \xmark & \xmark & \cmark & \cmark & Recon.\,+\,contr. \\
    MJEPA~\citep{teotia2026mjepa}
      & \cmark & \cmark & \cmark & \cmark & \xmark & \xmark & Latent pred. \\
    \specialrule{\lightrulewidth}{\aboverulesep}{0pt}
    \rowcolor{gray!12}
    \rule[-1.2ex]{0pt}{4ex}\textbf{\method (ours)}
      & \cmark & \cmark & \cmark & \cmark & \cmark & \cmark
      & \textbf{Invariance\,+\,SIGReg} \\
    \specialrule{\heavyrulewidth}{0pt}{0pt}
  \end{tabular}%
  }
  \endgroup
\end{table}

\paragraph{JEPAs} The Joint-Embedding Predictive Architecture~\citep{lecun2022path} predicts the embedding of one view from another instead of reconstructing inputs. I-JEPA~\citep{assran2023ijepa} instantiates it with spatial masking on images; V-JEPA~\citep{bardes2024vjepa} and V-JEPA\,2~\citep{assran2025vjepa2} extend it to video and world modelling; and LeJEPA~\citep{balestriero2025lejepa} supplies the theoretical grounding through SIGReg. MJEPA~\citep{teotia2026mjepa} is the closest audio-visual JEPA: it shares encoder parameters across modalities but trains with an EMA target encoder and a predictor network, both of which \method removes.\looseness-1

\paragraph{Cross-modal alignment and retrieval} CLIP~\citep{radford2021clip} and ALIGN~\citep{jia2021align} learn cross-modal similarity by contrasting matched and mismatched image--text pairs. AudioCLIP~\citep{guzhov2022audioclip}, ImageBind~\citep{girdhar2023imagebind}, and LanguageBind~\citep{zhu2024languagebind} extend shared embedding spaces to audio and further modalities. These approaches train explicit cross-modal alignment. In contrast, \method has no such objective: audio-only and video-only views are each pulled toward the same joint embedding, but are never compared directly or ranked against negatives. Retrieval therefore tests whether cross-modal correspondence emerges without being trained explicitly (\cref{sec:retrieval}).

\paragraph{Multi-view and modality-dropout strategies} SwAV~\citep{caron2020swav} introduced multi-crop training with few global and many local crops, a strategy inherited by DINO and LeJEPA. AV-HuBERT~\citep{shi2022avhubert} uses modality dropout for robustness to missing inputs. We bring these ideas together by making the missing modality define a partial view: sound alone and video alone must each approach a representation learned from their joint observation. The controlled ablation tests whether this change in view composition supplies a useful learning signal beyond ordinary cropping and masking.

\section{Methods}
\label{sec:method}
\method maps audio-only, video-only, and joint audio-visual inputs of the same clip to a common embedding. Each local view loses a modality and learns to match the joint audio-video centre. We first construct these views, then describe the shared encoder and the objective that brings their embeddings together while preventing collapse (\cref{fig:pipeline}).

\begin{figure}[t]
    \centering
    \setlength{\fboxsep}{1pt}
    \newcommand{\mycaption}[1]{\tikz[outer sep=0,inner sep=6pt]\node[minimum height=2.2cm,minimum width=3cm,text width=3cm,align=left,font=\scriptsize]{#1};}
    \newcommand{\mylabelR}[2]{\tikz[outer sep=0,inner sep=0]{\node[anchor=north west] (a) at (0,0) {#1}; \node[anchor=north west,fill=white,fill opacity=.5,inner sep=2pt,font=\scriptsize] at (a.north west) {#2};}}
    \newcommand{\mylabelL}[2]{\tikz[outer sep=0,inner sep=0]{\node[anchor=north east] (a) at (0,0) {#1}; \node[anchor=north east,fill=white,fill opacity=.5,inner sep=2pt,font=\scriptsize] at (a.north east) {#2};}}

    \begin{minipage}{\textwidth}
    \mylabelL{\includegraphics[height=2.2cm]{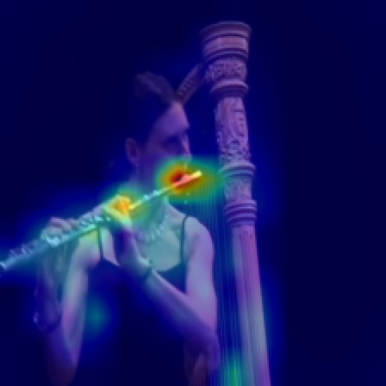}}{V}
    \hspace{0.01\linewidth}
    \mylabelR{\includegraphics[height=2.2cm]{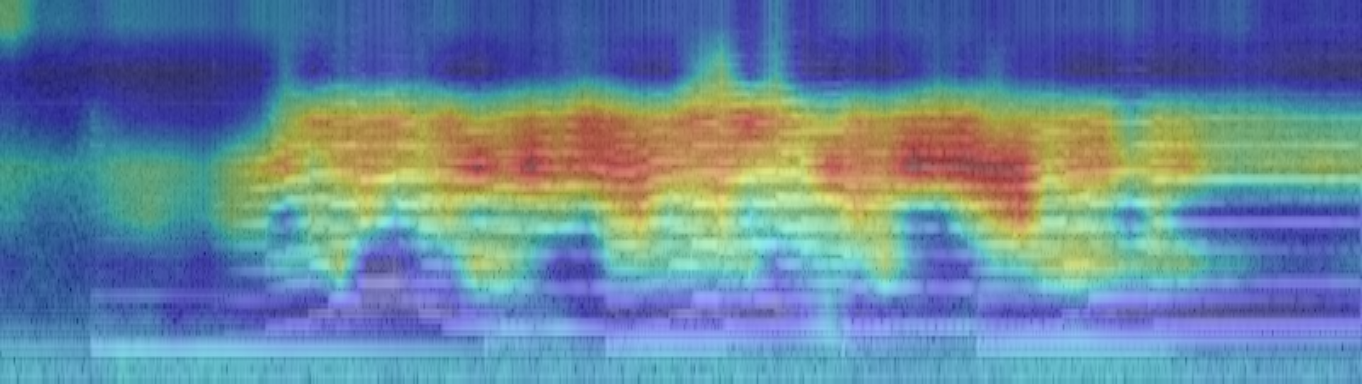}}{A}
    \mycaption{{\bf Flute:} The mouthpiece, where the sound is produced, is singled out rather than the whole instrument.}
    \end{minipage}

    \vspace*{.5em}

    \begin{minipage}{\textwidth}
    \mylabelL{\includegraphics[height=2.2cm]{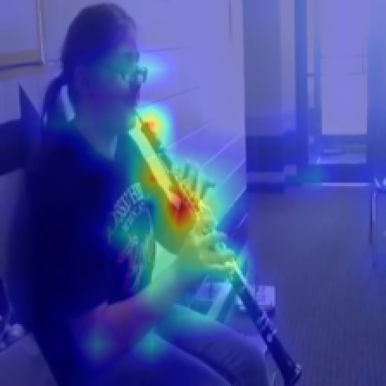}}{V}
    \hspace{0.01\linewidth}
    \mylabelR{\includegraphics[height=2.2cm]{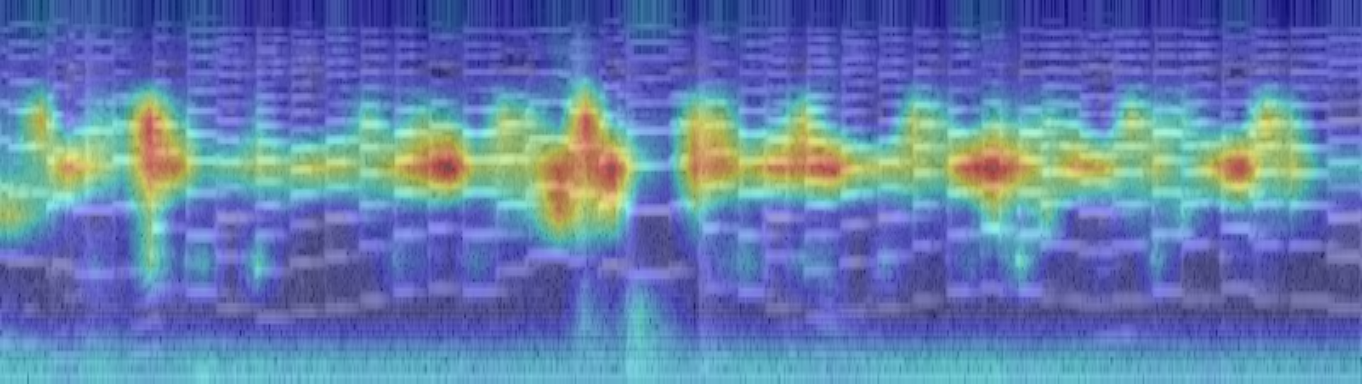}}{A}
    \mycaption{{\bf Oboe:} Attention falls on the person's hands and the part of the oboe they are playing.}
    \end{minipage}

    \vspace*{.5em}

    \begin{minipage}{\textwidth}
    \mylabelL{\includegraphics[height=2.2cm]{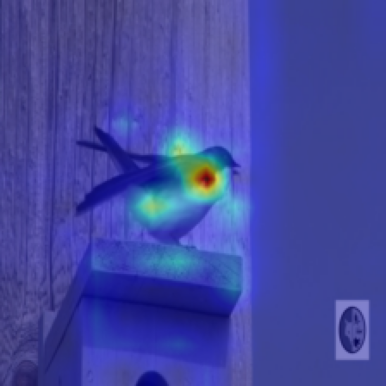}}{V}
    \hspace{0.01\linewidth}
    \mylabelR{\includegraphics[height=2.2cm]{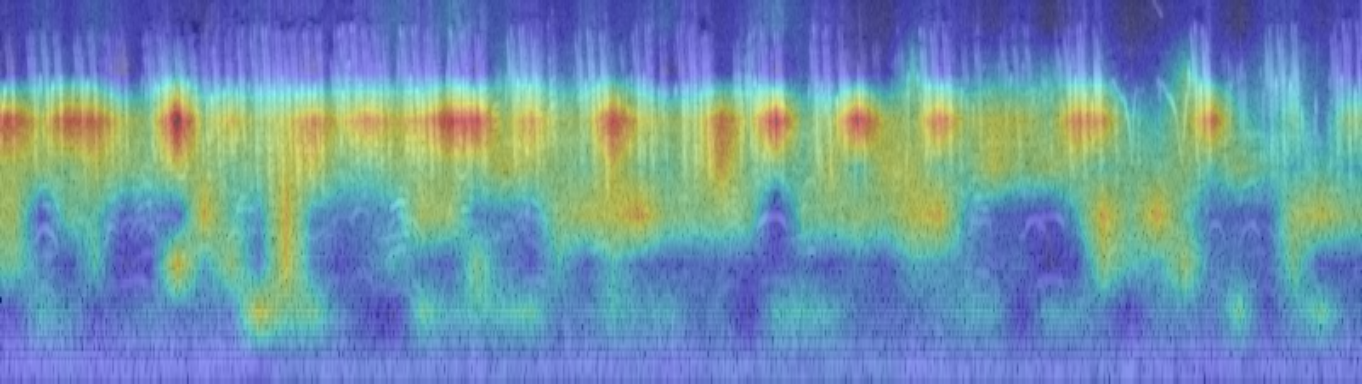}}{A}
    \mycaption{{\bf Bird:} A small bird stands out against a textured background, and each of its chirps is picked out in the spectrogram.}
    \end{minipage}
    \caption{
        \textbf{Attention finds the sound source without localization supervision.}
        Audio-to-video attention lands on the object producing the sound, and
        video-to-audio attention follows its harmonic and temporal structure.
        Neither behaviour is explicitly supervised; both emerge from matching
        single-modality views to a joint centre.
    }
    \label{fig:attention}
    \vspace*{-1em}
\end{figure}

\subsection{Constructing Joint and Partial Views}
\label{sec:cross_modal}
A joint view can use either sound or visual information to match another joint view. Removing one modality leaves the encoder with only the other, so an audio-only view must be matched from sound alone and a video-only view from appearance and motion alone. In both cases, the target is given by global views that retain both modalities.

\paragraph{Joint views define the centre}
We sample two 8-second temporal crops with random offsets from each 10-second clip. Both global views ($G{=}2$) contain audio and video, with a centre crop and random horizontal flip for video and a clean mel spectrogram for audio. After encoding and projection, we average their embeddings to form the joint centre $\bar{\vz}$. Each clip has its own centre, connecting observations of that clip.

\paragraph{Modality dropout creates the partial observations}
We also form two local views ($K{=}2$) from the temporal crops. Even-indexed local views retain only audio, with a clean spectrogram; odd-indexed views retain only video, with standard augmentations. We drop the absent modality's tokens entirely, shortening the sequence. Here, ``local'' refers to a partial set of modalities rather than a shorter temporal extent. This extends LeJEPA's spatial augmentations~\citep{balestriero2025lejepa} to the modality axis, so that each local view is a partial observation of the clip restricted to a single modality.

\paragraph{The target is a representation of the event}
The embeddings of both partial views are pulled toward the joint centre. Neither view has to reconstruct the missing modality, so the audio view does not predict visual motion and the video view does not predict the waveform. Each instead predicts an embedding computed from the joint observation, which encourages the encoder to retain information that audio and video share. \Cref{sec:view_ablation} tests this by varying only the local views.

\subsection{Encoding Joint and Partial Views with One Model}
\label{sec:architecture}
We project audio and video to tokens of the transformer's hidden dimension, so that a single transformer and projector process every view. Joint and partial views differ only in which tokens enter the sequence, and all are embedded in the same representation space.

\paragraph{Turning sound and video into tokens}
We resample audio to 16\,kHz and convert each 8-second crop into a mel spectrogram of shape $1\!\times\!128\!\times\!801$. A 2D convolution with kernel and stride $(16,16)$ produces $8\!\times\!50=400$ audio tokens. For video, a 3D convolution with kernel and stride $(2,16,16)$ maps $16$ RGB frames at $224\!\times\!224$ resolution to $8\!\times\!14^2=1568$ tubelet tokens. Factorized positional embeddings preserve frequency and time for audio, and spatial position and time for video. Learned modality-type embeddings identify which stream each token belongs to.

\paragraph{Sharing computation across views}
We prepend a learnable \texttt{[CLS]} token and concatenate the available tokens. A joint view has the sequence
\begin{equation}
[\texttt{CLS};\,\vv_1,\ldots,\vv_{1568};\,\va_1,\ldots,\va_{400}],
\end{equation}
while audio-only and video-only views contain 401 and 1569 tokens, respectively. Early fusion lets the transformer attend across modalities from its first layer when both are present. When one is absent, the same weights process the surviving tokens; no separate encoder or placeholder tokens for the missing modality are needed.

For ViT-B~\citep{dosovitskiy2021vit}, we use 12 layers, a hidden dimension of 768, and 12 attention heads; for ViT-L, we use 24 layers, a hidden dimension of 1024, and 16 attention heads. A shared three-layer MLP projects the \texttt{[CLS]} output through $d\!\to\!2048\!\to\!2048\!\to\!128$ dimensions, where $d$ is the hidden dimension, with BatchNorm and GELU. This projector maps every view into the space where we compute the loss; it is not a separate cross-modal predictor. \Cref{app:encoder_arch} and \cref{fig:encoder} give the full architecture.

\begin{figure}[!t]
  \centering
  \definecolor{audiocolor}{HTML}{D55E00}
  \definecolor{videocolor}{HTML}{2271B2}
  \definecolor{jointcolor}{HTML}{359B73}

  \newcommand{\audioicon}{%
    \tikz[baseline=-0.55ex]\fill[audiocolor] (0,0) circle (0.55ex);%
  }
  \newcommand{\videoicon}{%
    \tikz[baseline=-0.55ex]\fill[videocolor]
    (0,0.65ex) -- (-0.65ex,-0.5ex) -- (0.65ex,-0.5ex) -- cycle;%
  }
  \newcommand{\jointicon}{%
    \tikz[baseline=-0.55ex]\fill[jointcolor]
    (-0.55ex,-0.55ex) rectangle (0.55ex,0.55ex);%
  }
  \includegraphics[width=\linewidth]{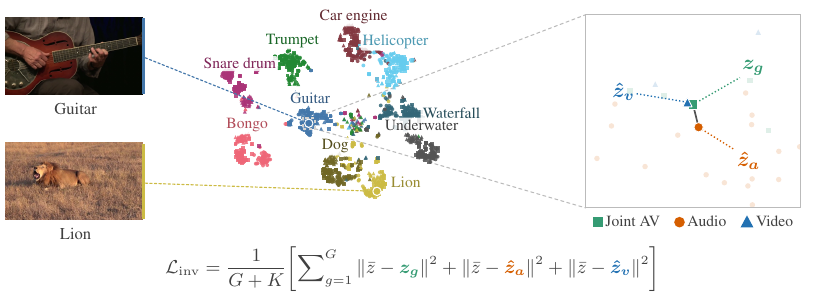}
  \caption{\textbf{Semantic classes cluster, and a clip's views stay close.} t-SNE of \texttt{[CLS]} embeddings, where each clip contributes an audio-only (\protect\audioicon), a video-only (\protect\videoicon), and a joint (\protect\jointicon) point. Within each cluster, the three shapes mix: audio-only and video-only views of a class land together even though they share no input. The zoom shows the guitar sample, whose joint embedding ($z_g$) lies between its audio ($\hat{z}_a$) and video ($\hat{z}_v$) embeddings, as the invariance loss (bottom) encourages.}
  \label{fig:alignment_semantic_pairs}
\end{figure}

\subsection{Matching Views without Collapse}
\label{sec:objective}
We want observations of the same clip to agree while retaining differences between clips. However, matching views alone would allow the encoder to return the same vector for every input. The LeJEPA objective~\citep{balestriero2025lejepa} addresses both requirements by combining an invariance loss with distributional regularization. Let $\vz_i=p_\phi(f_\theta(x_i))\in\mathbb{R}^{d_z}$ denote a projected view embedding and $\bar{\vz}=\frac{1}{G}\sum_{g=1}^{G}\vz_g$ the mean of the global embeddings for the same clip. We minimize
\begin{equation}
    \mathcal{L} = (1-\lambda)\,
                  \underbrace{\frac{1}{G+K}\sum_{i=1}^{G+K}
                  \|\bar{\vz} - \vz_i\|^2}_{\text{invariance}}
                  \;+\;\lambda\,
                  \underbrace{\vphantom{\sum_{i=1}^{G+K}}\text{SIGReg}(\{\vz_i\})}_{\text{regularization}},
    \label{eq:lejepa}
\end{equation}
where $\lambda$ balances agreement and distributional regularization. The invariance term is written for one clip and averaged over the minibatch; SIGReg regularizes embeddings across clips.

\paragraph{A learned target without a teacher}
The invariance term draws all views toward their clip's centre, and gradients also flow through the global embeddings that form it. The centre is therefore learned alongside the partial views, rather than being fixed or supplied by a second encoder. All views update the shared encoder and projector through the same loss, without a stop-gradient, EMA teacher, or cross-modal predictor. The centre can consequently adapt to information that the partial observations can express.

\paragraph{Preventing collapse with SIGReg}
SIGReg (\emph{Sketched Isotropic Gaussian Regularization}) supplies the pressure against collapse by matching the distribution of embeddings across clips to an isotropic Gaussian. It projects embeddings onto $M$ random unit directions and compares each projection's empirical characteristic function with the Gaussian target through the Epps--Pulley test. Resampling directions at each step constrains the distribution across different slices. For a fixed number of directions and evaluation points, the computation is linear in batch size and can aggregate the characteristic-function statistics across devices.

LeJEPA motivates the Gaussian target through its analysis of downstream probes: anisotropic covariance can amplify estimation error for linear probes, while an isotropic Gaussian minimizes integrated squared bias for the nonlinear probes considered in that analysis. These results motivate our regularizer under LeJEPA's assumptions; they do not guarantee optimal performance on arbitrary audio-visual tasks. The two terms act at different levels: invariance brings a clip's views together, while SIGReg preserves spread across clips. Their separate ablations in \cref{sec:view_ablation} test whether both are needed.

\paragraph{From a common target to cross-modal alignment}
No loss term compares audio-only and video-only embeddings directly. Both are instead pulled toward the same joint centre, and since $\|\vz_a-\vz_v\| \le \|\vz_a-\bar{\vz}\| + \|\bar{\vz}-\vz_v\|$, this also brings them close to each other. Retrieval additionally requires each embedding to lie closer to its partner than to other clips. SIGReg encourages this by spreading embeddings across clips, but the objective does not enforce it directly. \Cref{fig:alignment_semantic_pairs} illustrates the resulting neighbourhoods, and \cref{sec:retrieval} evaluates retrieval quantitatively.

\subsection{Monitoring the Representation}
\label{sec:probing}
\ifpreprint
We evaluate AudioSet-pretrained backbones with a linear probe on the \texttt{[CLS]} token and an attentive probe that pools patch tokens through one cross-attention query~\citep{bardes2024vjepa,teotia2026mjepa}. Since no probes are attached during pretraining, both are trained afterwards, either on the frozen backbone following standard audio-visual evaluation~\citep{gong2023cavmae,araujo2025cavmaesync} or jointly with the backbone during end-to-end fine-tuning. \Cref{app:training_configs} gives the training configurations.
\else
We monitor VGGSound features with a linear probe on the \texttt{[CLS]} token and an attentive probe that pools patch tokens through one cross-attention query~\citep{bardes2024vjepa,teotia2026mjepa}. During pretraining and fine-tuning, these probes receive detached features, so their losses cannot update the encoder. We fit AudioSet probes after pretraining with the backbone frozen, following standard audio-visual evaluation~\citep{gong2023cavmae,araujo2025cavmaesync}. \Cref{app:training_configs} gives the training configurations.
\fi

\section{Experiments}
\label{sec:experiments}
The evaluation first tests whether modality dropout is responsible for the improvement in representation quality (\cref{sec:view_ablation}). We then examine whether the learned features remain useful under frozen transfer (\cref{sec:exp_frozen}), align audio and video well enough for cross-modal retrieval (\cref{sec:retrieval}), and improve further with end-to-end fine-tuning (\cref{sec:exp_vggsound}).

\paragraph{Datasets and models}
AudioSet~\citep{gemmeke2017audioset} provides approximately two million clips with 527 multi-label sound classes; VGGSound~\citep{chen2020vggsound} provides approximately 184k training clips across 309 event classes. The ablations pretrain a ViT-B on VGGSound under a fixed short budget. The transfer, retrieval, and fine-tuning experiments use a ViT-B and a ViT-L pretrained on AudioSet-2M with two global and two modality-specific local views and $\lambda{=}0.05$. \Cref{app:training_configs} lists the full training configurations. All results are from single runs, except where noted.

\paragraph{Evaluation protocols}
Each table states whether it uses a frozen linear probe, a frozen attentive probe, or end-to-end fine-tuning. The probes measure information accessible in the pretrained features (\cref{sec:probing}); fine-tuning additionally lets supervised gradients adapt the backbone.

\subsection{Learning from Modality-Specific Partial Views}
\label{sec:view_ablation}
\begin{figure}[t]
\centering
\includegraphics[width=\linewidth]{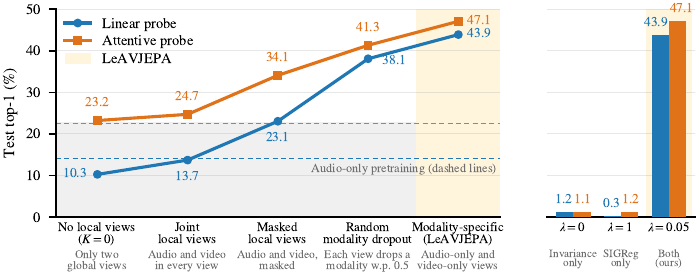}
\vspace*{-6pt}
\caption{\textbf{Modality-specific views improve cross-modal learning in \method.} Left: joint local views remain near the audio-only baselines (dashed lines). Random modality dropout improves probe accuracy, and making every local view audio-only or video-only yields the largest gain. Right: either loss term alone fails to learn useful features.}
\label{fig:components}
\end{figure}

The first test separates the contribution of the views from that of the encoder and loss. If ordinary joint multi-crop views were sufficient, the model could learn useful audio-visual features without ever losing a modality. We pretrain a ViT-B on VGGSound with the same encoder, hyperparameters, and training budget, varying the local-view construction and including a no-local-view control. \Cref{fig:components} compares the resulting online probe test accuracies; \cref{tab:views} in \cref{app:ablations} lists the values with top-5 accuracy, loss values, and embedding statistics.

\paragraph{Low loss does not imply useful cross-modal features}
Joint local views remain close to the audio-only baseline, while global views alone perform no better (\cref{fig:components}, left). The weak features do not arise from collapse: these variants achieve low invariance loss and near-unit embedding standard deviation (\cref{tab:views}). They instead reflect a limitation of the task, since when every view contains both modalities, the views can agree through one modality alone without the encoder relating it to the other. The results are consistent with this shortcut. Distributional regularity rules out collapse, but it does not show that the features combine information from audio and video.

\paragraph{Dropping a modality makes the task cross-modal}
When a local view contains only audio or only video, the encoder must match the joint centre from that modality alone. Random modality dropout improves both probes, and consistently making each local view audio-only or video-only yields the strongest features. The agreement between linear and attentive evaluation suggests that the benefit is not specific to one type of probe. Since these comparisons retain the same encoder, objective, and training budget, they identify view composition as a consequential design choice. Merely supplying both modalities does not confer the benefit obtained by learning from partial observations.

\paragraph{Both loss terms are necessary}
The loss ablations distinguish the roles of alignment and regularization (\cref{fig:components}, right). Removing SIGReg ($\lambda{=}0$) collapses the embedding and reduces classification toward chance. Removing invariance ($\lambda{=}1$) preserves embedding spread but also fails to learn useful features. Neither agreement alone nor distributional regularity alone is sufficient under this training protocol. The full objective makes the modality-specific task learnable while avoiding its trivial solution. Additional comparisons find similar performance with separate encoders and limited sensitivity to the tested SIGReg weights and local-view counts (\cref{app:ablations}), suggesting that the gain does not depend on one architectural configuration or a narrowly selected hyperparameter value.

\subsection{Frozen Transfer to AudioSet-20K and ESC-50}
\label{sec:exp_frozen}
The ablation shows that modality dropout produces better features on the pretraining dataset. We next test whether those features transfer to other tasks with the encoder frozen, so that downstream labels can train a probe but cannot change the representation. \Cref{tab:frozen} follows the frozen protocol of MJEPA~\citep{teotia2026mjepa}: an attentive probe on frozen AudioSet-pretrained features, evaluated on AudioSet-20K with audio-only (A), video-only (V), and audio-video (A-V) inputs, and on ESC-50~\citep{piczak2015esc}. Baseline rows are the re-evaluations reported by MJEPA under the same protocol.

\begin{table}[t]
\centering\footnotesize
\setlength{\abovecaptionskip}{0pt}
\setlength{\belowcaptionskip}{0pt}
\setlength{\tabcolsep}{5.5pt}
\caption{Frozen transfer with an attentive probe. AudioSet-20K mAP for audio-only (A), video-only (V), and audio-video (A-V) inputs, and ESC-50 accuracy. Baseline frozen results and parameter counts are as reported by \citet{teotia2026mjepa}\ifpreprint\else; the corresponding fully fine-tuned results are collected in \cref{tab:finetuned_reported} (\cref{app:finetuned})\fi. Best result per column in bold. $^{*}$Not reported in the MJEPA paper; provided by the authors through personal correspondence.}
\label{tab:frozen}
{\renewcommand{\arraystretch}{1.18}
\begin{tabular*}{\linewidth}{@{}l@{\hspace{4pt}}l@{\extracolsep{\fill}}llcccc@{}}
\toprule
& & & & \multicolumn{3}{c}{AudioSet-20K (mAP)} & ESC-50 \\
\cmidrule(lr){5-7}\cmidrule(lr){8-8}
& Method & Params & Pre-train data & A & V & A-V & Acc. \\
\midrule
\multirow{5}{*}[0pt]{\rotatebox[origin=c]{90}{\scriptsize Separate encoders}}
& CAV-MAE~\citep{gong2023cavmae}            & 170M & IN+AS & 19.38 & 18.14 & 34.59 & 77.5 \\
& MAViL~\citep{huang2023mavil}              & 170M & IN+AS & 30.00 & --    & --    & 90.8 \\
& EquiAV~\citep{kim2024equiav}              & 170M & IN+AS & 34.25 & 18.60 & 38.60 & 93.2 \\
& CAV-MAE Sync~\citep{araujo2025cavmaesync} & 170M & IN+AS & 21.66 & 16.20 & 28.50 & 89.2 \\
& XKD~\citep{sarkar2024xkd}                 & 170M & AS    & --    & --    & --    & 93.6 \\
\addlinespace[2pt]\midrule\addlinespace[2pt]
\multirow{4}{*}[0pt]{\rotatebox[origin=c]{90}{\scriptsize Shared encoder}}
& MJEPA ViT-B$^{*}$~\citep{teotia2026mjepa} & 86M  & AS    & 37.54 & 20.18 & 40.68 & -- \\
& MJEPA ViT-L~\citep{teotia2026mjepa}       & 300M & AS    & \textbf{38.89} & \textbf{25.38} & \textbf{42.90} & \textbf{95.2} \\
& \method ViT-B (ours)                     & 87M  & AS    & 27.79 & 14.82 & 30.98 & 88.0 \\
& \method ViT-L (ours)                     & 304M & AS    & 31.56 & 18.25 & 36.03 & 91.3 \\
\bottomrule
\end{tabular*}}
\end{table}

The frozen results show that both modalities contribute to the joint representation (\cref{tab:frozen}). Audio-only input is stronger than video-only input, yet combining them improves over audio alone for both model sizes. The larger backbone improves across audio, video, and joint evaluation, so its advantage is not restricted to a single input type. Transfer to ESC-50 further shows that features learned through the joint objective remain useful when only audio is available downstream.

\method is competitive with several separate-encoder methods, although both model sizes remain below MJEPA. This supports the utility of the simpler objective, which reaches these results without MJEPA's EMA teacher or predictor. The probe follows the evaluation protocol of MJEPA~\citep{teotia2026mjepa}\ifpreprint, with a four-layer attentive probe trained for 40 epochs\else, and \cref{app:audioset_probe_fits} documents its settings\fi.

\subsection{Cross-Modal Retrieval without a Contrastive Loss}
\label{sec:retrieval}
Classification can succeed without placing paired audio and video close enough for retrieval. We therefore test a more specific consequence of the common-centre objective: can a single-modality embedding identify its partner among other clips? We encode each clip twice, from its audio tokens alone and from its video tokens alone, and rank candidates by cosine similarity of the projected \texttt{[CLS]} features. We evaluate on the balanced AudioSet retrieval subset released with CAV-MAE~\citep{gong2023cavmae}, the pool also used by CAV-MAE Sync, EquiAV, and MJEPA.

\begin{table}[t]
\centering\small
\setlength{\abovecaptionskip}{0pt}
\setlength{\belowcaptionskip}{0pt}
\caption{AudioSet cross-modal retrieval on the CAV-MAE balanced evaluation subset (1,725 clips; 1,722 available for our evaluation), recall in percent. Baseline rows are as reported in the respective papers and all train an explicit cross-modal alignment term or predictor; \method{} rows are zero-shot cosine retrieval unless marked. Best baseline result per column in bold. $^{\dagger}$MJEPA retrieves through its jointly trained cross-modal predictor with an L1 nearest-neighbour search rather than direct embedding similarity. $^{\ddagger}$Contrastive alignment head trained on frozen features; not zero-shot.}
\label{tab:retrieval}
\begin{tabular*}{\linewidth}{@{\extracolsep{\fill}}lrrrrrr@{}}
\toprule
& \multicolumn{3}{c}{Audio $\to$ video} & \multicolumn{3}{c}{Video $\to$ audio} \\
\cmidrule(lr){2-4}\cmidrule(lr){5-7}
Method & R@1 & R@5 & R@10 & R@1 & R@5 & R@10 \\
\midrule
CAV-MAE~\citep{gong2023cavmae}            & 15.1 & 34.0 & 43.0 & 18.8 & 39.5 & 50.1 \\
CAV-MAE Sync~\citep{araujo2025cavmaesync} & 27.9 & 52.4 & 62.2 & \textbf{35.2} & \textbf{58.3} & \textbf{67.6} \\
EquiAV~\citep{kim2024equiav}              & \textbf{29.6} & \textbf{53.7} & \textbf{63.1} & 30.1 & 53.3 & 62.9 \\
MJEPA ViT-L$^{\dagger}$~\citep{teotia2026mjepa} & 29.3 & \textbf{53.7} & 61.8 & 30.6 & 53.4 & 61.3 \\
\midrule
\method{} ViT-B (ours)                   & 10.5 & 28.9 & 36.6 & 11.4 & 28.2 & 36.8 \\
\method{} ViT-L (ours)                   & 19.1 & 40.5 & 50.4 & 18.7 & 42.1 & 51.9 \\
\method{} ViT-L + InfoNCE head$^{\ddagger}$ (ours) & 24.4 & 46.9 & 57.7 & 25.6 & 48.0 & 57.1 \\
\bottomrule
\end{tabular*}
\end{table}

Direct cosine retrieval is well above chance in both directions and improves consistently with the larger backbone (\cref{tab:retrieval}). This shows that the model can identify a paired observation even when the query and candidate share no input modality. Nevertheless, the strongest methods trained with explicit cross-modal alignment remain ahead, indicating that matching a joint centre leaves room to improve cross-modal alignment.

To test whether the frozen features support stronger retrieval with a learned head, we fit a contrastive (InfoNCE) head on 300 AudioSet training clips per class and select the best head on a held-out validation set. The head improves retrieval in both directions without updating the encoder.

\subsection{End-to-End Fine-Tuning on VGGSound}
\label{sec:exp_vggsound}
\ifpreprint\else
We finally allow supervision to adapt the entire encoder, testing downstream performance under the end-to-end protocol used by most published audio-visual SSL comparisons. \Cref{app:finetuned} documents the training settings and an additional comparison with CAV-MAE Sync under a matched frozen probe.
\fi

\begin{table}[t]
\centering\footnotesize
\setlength{\abovecaptionskip}{0pt}
\setlength{\belowcaptionskip}{0pt}
\caption{Audio-visual classification on VGGSound (top-1, end-to-end fine-tuning). Published baselines all use separate audio and video encoders, except AVSiam, and are listed above the rule; our AudioSet-pretrained shared-encoder ViT-B (87M) and ViT-L (304M) below it. Best result per block in bold.}
\label{tab:main}
\setlength{\tabcolsep}{18pt}
\begin{tabular}{@{}lrllc@{}}
\toprule
Method & Params & Init. & Pretraining & Top-1 \\
\midrule
AV-MAE~\citep{georgescu2023avmae} & 172M & -- & VGGS & 64.2 \\
AV-MAE~\citep{georgescu2023avmae} & 611M & -- & VGGS & 65.0 \\
AVSiam~\citep{lin2024avsiam} & 100M & IN-21K SL & AS-2M & 64.9 \\
AVSiam~\citep{lin2024avsiam} & 332M & IN-21K SL & AS-2M & 67.1 \\
CAV-MAE~\citep{gong2023cavmae} & 164M & IN-SSL & AS-2M & 65.5 \\
MAViL~\citep{huang2023mavil} & 172M & IN-SSL & AS-2M & \textbf{67.1} \\
EquiAV~\citep{kim2024equiav} & 172M & IN-SSL & AS-2M & \textbf{67.1} \\
\midrule
\method (Linear) & 87M & -- & AS-2M & 58.3 \\
\method (Attentive) & 87M & -- & AS-2M & 60.4 \\
\method (Linear) & 304M & -- & AS-2M & 59.9 \\
\method (Attentive) & 304M & -- & AS-2M & \textbf{61.1} \\
\bottomrule
\end{tabular}
\end{table}

\ifpreprint We next fine-tune the entire encoder end to end, as in most published audio-visual SSL comparisons. \fi Attentive pooling improves over the linear head for both model sizes (\cref{tab:main}), showing that the choice of classification head still matters after fine-tuning. However, the advantage of the larger backbone is less pronounced here than in frozen transfer or retrieval. This contrast shows why frozen and fine-tuned performance should be reported separately, since fine-tuning can compensate for differences in representation quality that frozen evaluation exposes.

The published baselines retain an advantage after fine-tuning, with a similar ordering on AudioSet-20K\ifpreprint\else{} (\cref{tab:finetuned_reported})\fi. Differences in initialization, objectives, and optimization prevent attributing this gap to any individual component. The shared-encoder approach therefore supports useful transfer, while matching the strongest fully adapted systems remains an open challenge.

\section{Discussion and Conclusion}
\label{sec:conclusion}
\method shows that the construction of partial views is central to learning useful audio-visual representations. Treating sound and vision as incomplete observations of the same event allows a shared encoder to learn cross-modal correspondence through a symmetric objective. This extends the role of view construction in self-supervised learning, since withholding a modality determines which input the encoder must use to predict a representation learned from both modalities. Agreement between views is informative only insofar as the views define a useful learning task. Modality-specific observations substantially improve the learned features under the same encoder and loss, demonstrating that the information retained in each view deserves explicit consideration when designing multimodal objectives. The contribution of \method is to establish that useful alignment and transfer can already emerge without a dedicated cross-modal predictor, reconstruction decoder, or teacher network. More broadly, our results show that a single encoder, a symmetric objective, and views that each contain only one modality are enough to learn strong cross-modal representations, offering a simple, scalable foundation for future multimodal models.

\paragraph{Limitations and future work}
Our evaluation relies mainly on sound-centric benchmarks, leaving the encoder's broader visual capabilities unresolved. Published baselines also use different training protocols, so performance gaps are hard to attribute. Moreover, matching partial observations assumes that they contain sufficient shared information. Off-screen sounds and unrelated visual activity offer relevant tests of when that assumption no longer holds. A natural next step is to extend modality dropout beyond audio and video, for example to text or depth, and to evaluate on video-centric tasks. With the method reduced to one encoder and two loss terms, another is to make it cheaper to train, for example with fewer views, tokens, or steps.

\ifpreprint
\subsubsection*{Acknowledgments}
We acknowledge the computational resources provided by the Aalto Science-IT project (Triton cluster) and CSC -- IT Center for Science, Finland.
\fi

\clearpage

\ifpreprint\else
\subsection*{AI use statement}
We used generative AI tools to edit the text of this paper, including grammar, phrasing, and condensing prose; to write and debug parts of the training, evaluation, and plotting code, and to set up and manage experiments; to assist with parts of the experimental design; and to help search for and summarise some of the related work. We did not use generative AI tools to generate the core research idea or method. We have reviewed all AI-assisted work. AI-generated code and experiment configurations were verified and tested by the authors before use in any reported experiment, and the authors made all final decisions on experimental design. We take responsibility for the final content of this work, including text, claims or artifacts produced with the aid of generative AI.

\subsection*{Ethics statement}
This work uses only publicly available research datasets (AudioSet, VGGSound, and ESC-50) under their existing terms of use. It involves no human subjects, no new data collection, and no release of new data. The authors declare no conflicts of interest.

\subsection*{Reproducibility statement}
Anonymous source code is provided as supplementary material, covering the model, data pipeline, pretraining, fine-tuning, probing, and retrieval, together with the job scripts for the reported runs. The code and pretrained checkpoints will be released publicly upon acceptance. \Cref{sec:method} describes the view construction, encoder, and objective, and \cref{app:method_details} specifies the encoder architecture and probe readouts. \Cref{app:training_configs,app:finetune_protocol,app:probe_training} list the full pretraining, fine-tuning, and probe configurations, including optimiser settings, batch sizes, and hardware. All experiments use public datasets.
\fi

\bibliographystyle{iclr2027_conference}

\addtocontents{toc}{\protect\setcounter{tocdepth}{3}}

\clearpage
\appendix

\raggedbottom
\setlength{\intextsep}{12pt plus 2pt minus 2pt}
\captionsetup{font=small,labelfont=bf,skip=7pt,
  justification=raggedright,singlelinecheck=false}
\captionsetup[subfigure]{font=small,labelfont=normalfont,
  justification=centering,singlelinecheck=true,skip=4pt}
\makeatletter
\renewcommand\paragraph{\@startsection{paragraph}{4}{\z@}%
  {1.2ex \@plus .2ex \@minus .1ex}{-1em}%
  {\normalfont\normalsize\bfseries}}
\makeatother

\setcounter{table}{0}
\setcounter{figure}{0}
\renewcommand{\thetable}{A\arabic{table}}
\renewcommand{\thefigure}{A\arabic{figure}}

\section*{Appendices}

\Cref{app:method_details} describes the encoder and the linear and attentive
probes. \Cref{app:implementation_details} documents pretraining, fine-tuning,
and probe training. \Cref{app:additional_results} presents training curves,
qualitative analyses of the learned representations, additional benchmark
comparisons, and ablation and hyperparameter studies.

\begingroup
\hypersetup{linkcolor=black}
\makeatletter
\renewcommand{\l@subsubsection}{\@dottedtocline{3}{3.8em}{3.4em}}
\@starttoc{toc}
\makeatother
\endgroup
\clearpage

\section{Method Details}
\label[appendix]{app:method_details}

\subsection{Encoder Architecture}
\label{app:encoder_arch}

\begin{figure}[H]
  \centering
  \includegraphics[width=\textwidth]{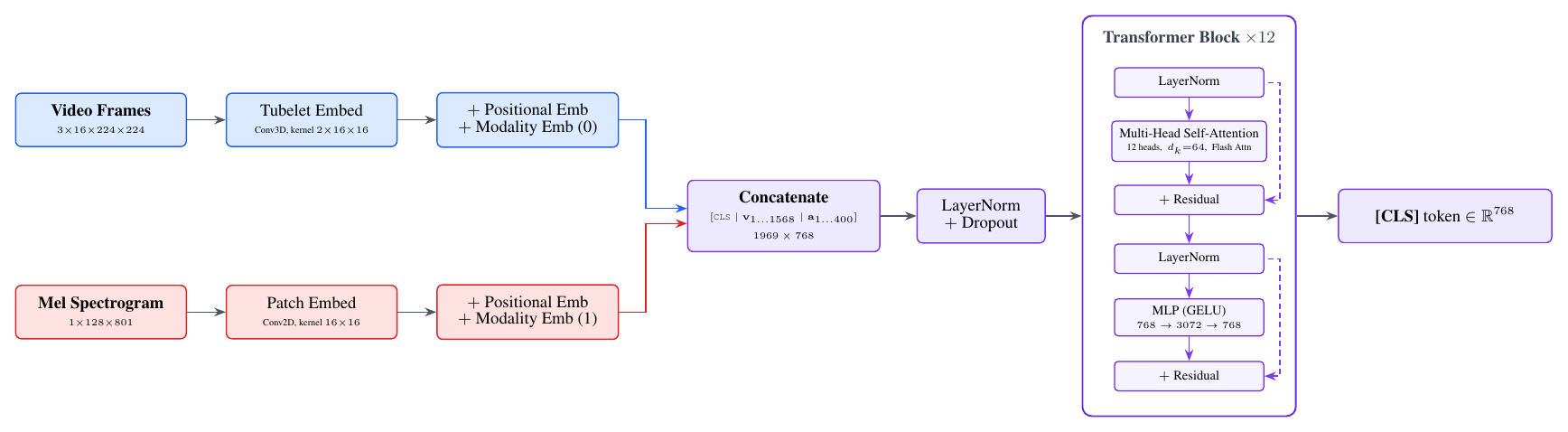}
  \caption{\textbf{\method early-fusion encoder.} Video tubelets and
  mel-spectrogram patches are embedded, summed with factorized positional and
  modality-type embeddings, concatenated with a \texttt{[CLS]} token, and
  processed by a $12$-layer ViT-Base. The same encoder is used for every view,
  including the modality-dropout local views, where the absent modality
  contributes no tokens and the sequence is correspondingly shorter.}
  \label{fig:encoder}
\end{figure}

\Cref{fig:encoder} shows the full \method encoder used for every view (global
and local) during pretraining. Video frames of shape
$3{\times}16{\times}224{\times}224$ are tokenized by a $2{\times}16{\times}16$
Conv3D tubelet into $1568$ video tokens; the mel spectrogram of shape
$1{\times}128{\times}801$ is tokenized by a $16{\times}16$ Conv2D into $400$
audio tokens. Both streams receive factorized positional embeddings (spatial +
temporal for video; frequency + time for audio) and a learnable modality-type
embedding (ID 0 for video, ID 1 for audio). A learnable \texttt{[CLS]} token is
prepended and the resulting $1969$-token sequence is passed through a $12$-layer
ViT-Base ($d{=}768$, $12$ heads, FlashAttention-2). On local views, the tokens of the
dropped modality are omitted from the sequence entirely (\texttt{[CLS]} plus
$1568$ video or $400$ audio tokens), so the same encoder handles joint,
audio-only, and video-only inputs without architectural changes.
The \texttt{[CLS]} output is taken as the clip embedding and fed to the
projection MLP for the LeJEPA loss (\cref{fig:pipeline}).

\subsection{Linear and Attentive Probes}
\label{app:readouts}

The linear probe applies LayerNorm and a linear classifier to the \texttt{[CLS]} token. The attentive probe instead pools encoder patch tokens through cross-attention. Concretely, the attentive probe maintains a single learnable query that attends to the full sequence of encoder patch tokens (audio and video, excluding \texttt{[CLS]}) through a multi-head cross-attention layer ($d{=}768$, 12 heads), followed by LayerNorm and a linear classifier that produces the class logits. Unlike the linear probe, which is forced to summarize the sequence through the \texttt{[CLS]} token alone, the attentive probe adaptively aggregates the most informative audio and video patches per sample, providing a more expressive probe of the frozen backbone~\citep{bardes2024vjepa,teotia2026mjepa}.

\clearpage

\section{Implementation Details}
\label[appendix]{app:implementation_details}

\label[appendix]{app:training_configs}

\subsection{Pretraining Configuration}

\Cref{tab:pretrain_config} lists the AudioSet-2M pretraining configurations of the ViT-B and ViT-L backbones. No online probes are attached during AudioSet pretraining; AudioSet probes are fitted post hoc on the frozen backbone (\cref{app:probe_training}).

\begin{table}[H]
\caption{\textbf{AudioSet-2M pretraining configurations.} Settings spanning both columns are shared.}
\label{tab:pretrain_config}
\centering\small
\setlength{\tabcolsep}{5pt}
\renewcommand{\arraystretch}{1.06}
\newcommand{\sharedsetting}[1]{\multicolumn{2}{l@{}}{#1}}
\setlength{\belowrulesep}{0pt}
\newcommand{\configgroup}[1]{\multicolumn{3}{@{}>{\columncolor{black!5}[0pt][0pt]}l@{}}{\rule[-0.9ex]{0pt}{3.1ex}\textbf{#1}}\\}
\begin{tabularx}{\linewidth}{@{}p{0.30\linewidth}XX@{}}
\toprule
\rule{0pt}{2.8ex}\textbf{Setting} & \textbf{ViT-B} & \textbf{ViT-L} \\
\midrule
\configgroup{Data}
Dataset                  & \sharedsetting{AudioSet-2M (unbalanced), ${\sim}$1.9M clips} \\
Clip length              & \sharedsetting{8\,s (from a 10\,s clip, random offset)} \\
\midrule
\configgroup{Architecture}
Encoder                  & 12L, 768d, 12h & 24L, 1024d, 16h \\
Parameters               & 87M & 304M \\
Audio embed              & \sharedsetting{Conv2D, kernel/stride $16{\times}16$} \\
Video embed              & \sharedsetting{Conv3D, kernel/stride $2{\times}16{\times}16$} \\
Sequence length          & \sharedsetting{1969 (1 CLS + 1568 video + 400 audio)} \\
Projector                & $768{\to}2048{\to}2048{\to}128$ & $1024{\to}2048{\to}2048{\to}128$ \\
\midrule
\configgroup{LeJEPA training procedure}
Global views $G$         & \sharedsetting{2} \\
Local views $K$          & \sharedsetting{2 (1 audio-only, 1 video-only)} \\
$\lambda$ (SIGReg weight) & \sharedsetting{0.05} \\
\midrule
\configgroup{Optimization}
Optimizer                & \sharedsetting{AdamW~\citep{loshchilov2019adamw}} \\
Learning rate            & \sharedsetting{$5{\times}10^{-4}$} \\
Weight decay             & \sharedsetting{0.05} \\
LR schedule              & \sharedsetting{Linear warmup (15\%) + cosine to $10^{-6}$} \\
Gradient clipping        & \sharedsetting{5.0} \\
Mixed precision          & \sharedsetting{bf16} \\
Gradient checkpointing   & No & Yes \\
Batch size (per GPU)     & 40 & 128 \\
Effective batch size     & 320 & 1024 \\
\midrule
\configgroup{Compute}
GPUs                     & 8$\times$ NVIDIA H200 & 8$\times$ NVIDIA B300 \\
Parallelism              & \sharedsetting{DDP} \\
Wall-clock time          & ${\sim}192$\,h (to epoch 57) & ${\sim}160$\,h (to epoch 66) \\
\bottomrule
\end{tabularx}
\end{table}

\clearpage

\subsection{Fine-Tuning Configuration}
\label{app:finetune_protocol}

\Cref{tab:finetune_config} lists the VGGSound fine-tuning configuration used for both AudioSet-pretrained backbones in \cref{tab:main}. The two runs share the recipe and effective batch size.

\begin{table}[H]
\caption{\textbf{VGGSound fine-tuning configuration} for the AudioSet-pretrained ViT-B and ViT-L backbones. Settings spanning both columns are shared.}
\label{tab:finetune_config}
\centering\small
\renewcommand{\arraystretch}{1.08}
\setlength{\tabcolsep}{5pt}
\newcommand{\sharedsetting}[1]{\multicolumn{2}{l@{}}{#1}}
\setlength{\belowrulesep}{0pt}
\newcommand{\configgroup}[1]{\multicolumn{3}{@{}>{\columncolor{black!5}[0pt][0pt]}l@{}}{\rule[-0.9ex]{0pt}{3.1ex}\textbf{#1}}\\}
\begin{tabularx}{\linewidth}{@{}p{0.30\linewidth}XX@{}}
\toprule
\rule{0pt}{2.8ex}\textbf{Setting} & \textbf{ViT-B} & \textbf{ViT-L} \\
\midrule
\configgroup{Probes on top of pretrained backbone}
Linear classifier        & \sharedsetting{LayerNorm + linear on \texttt{[CLS]} (309 classes)} \\
Attentive classifier     & 1 query, 12-head cross-attention & 1 query, 16-head cross-attention \\
\midrule
\configgroup{Optimization}
Optimizer                & \sharedsetting{AdamW} \\
Head learning rate       & \sharedsetting{$2{\times}10^{-4}$} \\
Backbone learning rate   & \sharedsetting{$2{\times}10^{-5}$ ($0.1\times$ head LR)} \\
Weight decay             & \sharedsetting{0.05} \\
LR schedule              & \sharedsetting{Linear warmup (5\%) + cosine to $10^{-7}$} \\
Gradient clipping        & \sharedsetting{1.0} \\
Label smoothing          & \sharedsetting{0.1} \\
Mixed precision          & \sharedsetting{bf16} \\
Batch size (per GPU)     & 80 & 40 \\
Effective batch size     & \sharedsetting{320} \\
Epochs                   & \sharedsetting{20} \\
\midrule
\configgroup{Data and augmentation}
Views                    & \sharedsetting{Single clip, no SSL views or masking} \\
SpecAugment              & \sharedsetting{Frequency mask 24, time mask 80} \\
Mixup $\alpha$           & \sharedsetting{0.2} \\
Random resized crop      & \sharedsetting{Min.\ scale 0.5} \\
\midrule
\configgroup{Compute}
GPUs                     & 4$\times$ NVIDIA H200 & 8$\times$ NVIDIA H200 \\
Parallelism              & \sharedsetting{DDP} \\
\bottomrule
\end{tabularx}
\end{table}

\subsection{Probe Training and Evaluation}
\label{app:probe_training}

\ifpreprint
Since no probes are attached during AudioSet pretraining, the backbone is evaluated afterwards with two classification probes: a \textbf{linear probe} (LayerNorm + linear classifier on the \texttt{[CLS]} token) and an \textbf{attentive probe} that pools information across patch tokens via cross-attention. For frozen evaluation (\cref{sec:exp_frozen}), both probes are trained post-hoc on top of the frozen AudioSet-pretrained backbone, as is standard practice in audio-visual SSL~\citep{gong2023cavmae,araujo2025cavmaesync}, whereas for fine-tuning (\cref{sec:exp_vggsound}) they are trained jointly with the backbone using the settings in \cref{tab:finetune_config}.
\else
During VGGSound pretraining and fine-tuning, two classification probes are trained on frozen backbone features (gradients detached from the encoder): a \textbf{linear probe} (LayerNorm + linear classifier on the \texttt{[CLS]} token) and an \textbf{attentive probe} that pools information across patch tokens via cross-attention. We do not attach these probes during AudioSet pretraining; AudioSet probes used at evaluation time are trained post-hoc on top of the frozen AudioSet-pretrained backbone (\cref{sec:exp_frozen}). Both probes are trained with cross-entropy loss (label smoothing 0.1) using a separate learning rate ($10^{-3}$) and no weight decay, following the LeJEPA evaluation protocol of frozen-backbone probing, which is also standard practice in audio-visual SSL~\citep{gong2023cavmae,araujo2025cavmaesync}.
\fi

\clearpage

\section{Additional Results}
\label[appendix]{app:additional_results}

\subsection{Training Curves}
\label{app:training_dynamics}

\subsubsection{Embedding-Quality Curves}
\label{app:embed_std}

\begin{figure}[H]
  \centering
  \begin{subfigure}[t]{0.49\linewidth}
    \centering
    \includegraphics[width=0.8\linewidth]{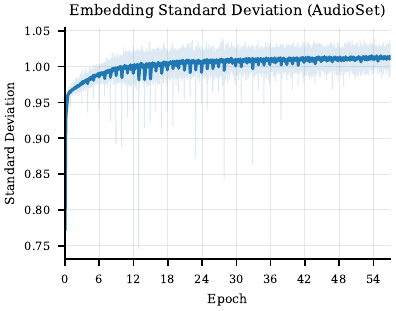}
    \caption{AudioSet-2M (57 ep.)}
  \end{subfigure}\hfill
  \begin{subfigure}[t]{0.49\linewidth}
    \centering
    \includegraphics[width=0.8\linewidth]{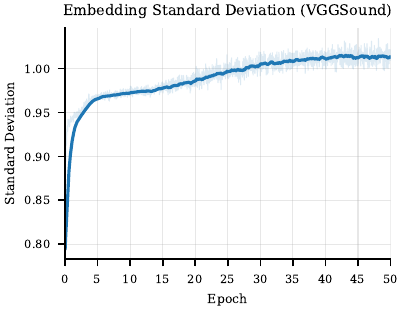}
    \caption{VGGSound (50 ep.)}
  \end{subfigure}
  \caption{\textbf{Embedding standard deviation} over training. SIGReg drives
  the per-dimension std toward 1.0 (the isotropic Gaussian target).}
  \label{fig:embed_std}
\end{figure}

\cref{fig:embed_std} tracks the per-dimension standard deviation of the
projected embeddings over training for both pretraining runs. In each case the
value rises rapidly toward ${\sim}1.01$ (from ${\sim}0.8$ on VGGSound and
${\sim}0.8$ on AudioSet), confirming that SIGReg drives the embedding
distribution toward the target isotropic Gaussian $\mathcal{N}(\vzero, \mI)$ at
both VGGSound and AudioSet scales.

\subsubsection{VGGSound Pretraining Curves}
\label{app:curves}

\begin{figure}[H]
\centering
    \begin{subfigure}[t]{0.49\linewidth}
        \centering
        \includegraphics[width=0.9\linewidth]{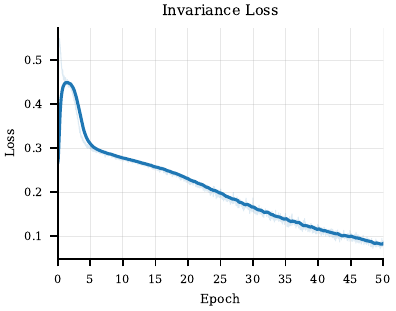}
        \caption{Invariance loss}
        \label{fig:inv_loss}
    \end{subfigure}
    \hfill
    \begin{subfigure}[t]{0.49\linewidth}
        \centering
        \includegraphics[width=0.9\linewidth]{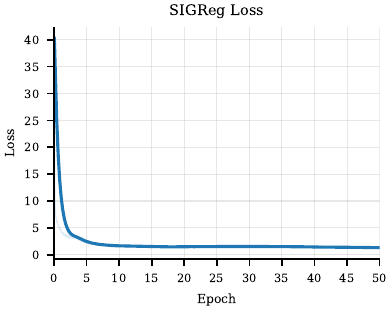}
        \caption{SIGReg loss}
        \label{fig:sigreg_loss}
    \end{subfigure}
    \caption{
        \textbf{Pretraining loss components} (VGGSound-only run, 50 epochs).
    }
    \label{fig:loss_components}
\end{figure}

\paragraph{Pretraining loss components}
\cref{fig:loss_components} decomposes the LeJEPA objective into its
two unweighted terms.
The invariance term decreases steadily, indicating that view
embeddings converge toward the global center $\bar{\vz}$ and that
cross-modal alignment is successful.
The SIGReg term drops sharply during the first epoch and then
plateaus near zero, confirming that the embedding distribution
matches the target isotropic Gaussian early in training and remains
there for the rest of the run.

\clearpage

\begingroup
\setlength{\intextsep}{7pt}
\captionsetup{skip=5pt}
\paragraph{Online probe curves}
\cref{fig:probe_curves} reports the online linear and attentive
probes on frozen backbone features during pretraining: attentive
probe top-1 accuracy, top-5 accuracy for both probes, and the
combined cross-entropy loss.
The attentive probe consistently outperforms the linear probe by
${\sim}2$--3 points across both top-1 and top-5.

\begin{figure}[H]
    \begin{subfigure}[t]{0.32\linewidth}
        \centering
        \includegraphics[width=\linewidth]{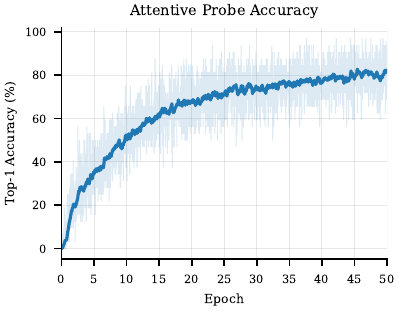}
        \caption{Attentive probe top-1}
        \label{fig:att_probe_acc}
    \end{subfigure}
    \hfill
    \begin{subfigure}[t]{0.32\linewidth}
        \centering
        \includegraphics[width=\linewidth]{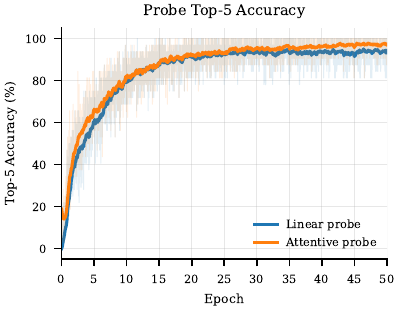}
        \caption{Linear / attentive top-5}
        \label{fig:probe_top5}
    \end{subfigure}
    \hfill
    \begin{subfigure}[t]{0.32\linewidth}
        \centering
        \includegraphics[width=\linewidth]{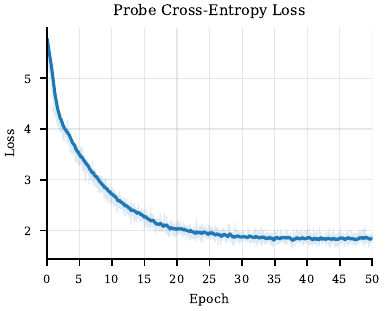}
        \caption{Probe cross-entropy loss}
        \label{fig:probe_loss}
    \end{subfigure}
    \caption{
        \textbf{Online probing curves} on frozen backbone features
        during pretraining.
    }
    \label{fig:probe_curves}
\end{figure}

\subsubsection{AudioSet Pretraining Curves}
\label{app:audioset_curves}

We complement the AudioSet pretraining loss in the main text with the
per-component invariance and SIGReg curves
(\cref{fig:loss_components_audioset}).
The qualitative behaviour mirrors the VGGSound-only run: the invariance term
decreases steadily and the SIGReg term drops sharply during the first epoch and
then plateaus near zero, confirming that SIGReg behaves as expected at AudioSet
scale.

\begin{figure}[H]
\centering
    \begin{subfigure}[t]{0.49\linewidth}
        \centering
        \includegraphics[width=0.78\linewidth]{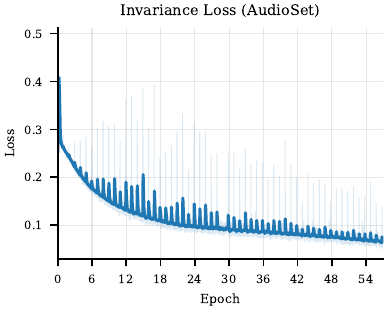}
        \caption{Invariance loss}
        \label{fig:inv_loss_audioset}
    \end{subfigure}
    \hfill
    \begin{subfigure}[t]{0.49\linewidth}
        \centering
        \includegraphics[width=0.78\linewidth]{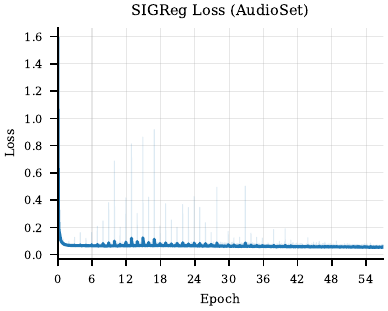}
        \caption{SIGReg loss}
        \label{fig:sigreg_loss_audioset}
    \end{subfigure}
    \caption{
        \textbf{AudioSet pretraining loss components} over 57 epochs of
        AudioSet-2M.
    }
    \label{fig:loss_components_audioset}
\end{figure}

\subsubsection{Fine-Tuning Curves}
\label{app:finetune}

\cref{fig:ft_curves} shows the end-to-end fine-tuning curves
(top-1/top-5 accuracy and cross-entropy loss) when the pretrained
\method backbone is fine-tuned with a linear classifier on the
\texttt{[CLS]} token.

\begin{figure}[H]
\centering
    \begin{subfigure}[t]{0.49\linewidth}
        \centering
        \includegraphics[width=0.78\linewidth]{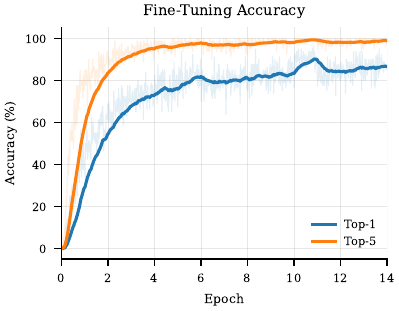}
        \caption{Top-1 / Top-5 accuracy}
        \label{fig:ft_acc}
    \end{subfigure}
    \hfill
    \begin{subfigure}[t]{0.49\linewidth}
        \centering
        \includegraphics[width=0.78\linewidth]{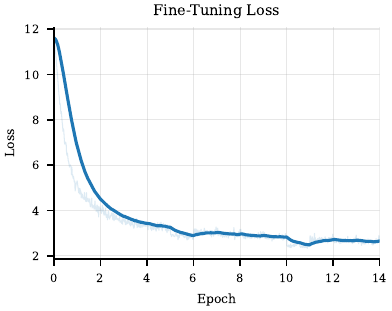}
        \caption{Cross-entropy loss}
        \label{fig:ft_loss}
    \end{subfigure}
    \caption{
        \textbf{Fine-tuning curves} for end-to-end fine-tuning of the
        pretrained \method backbone.
    }
    \label{fig:ft_curves}
\end{figure}

\endgroup

\clearpage

\subsection{Qualitative Analysis of Learned Representations}
\label{app:qualitative}

\subsubsection{Semantic Structure of the Embedding Space}
\label{app:tsne_family}

\begin{figure}[H]
  \centering
  \includegraphics[width=0.72\linewidth]{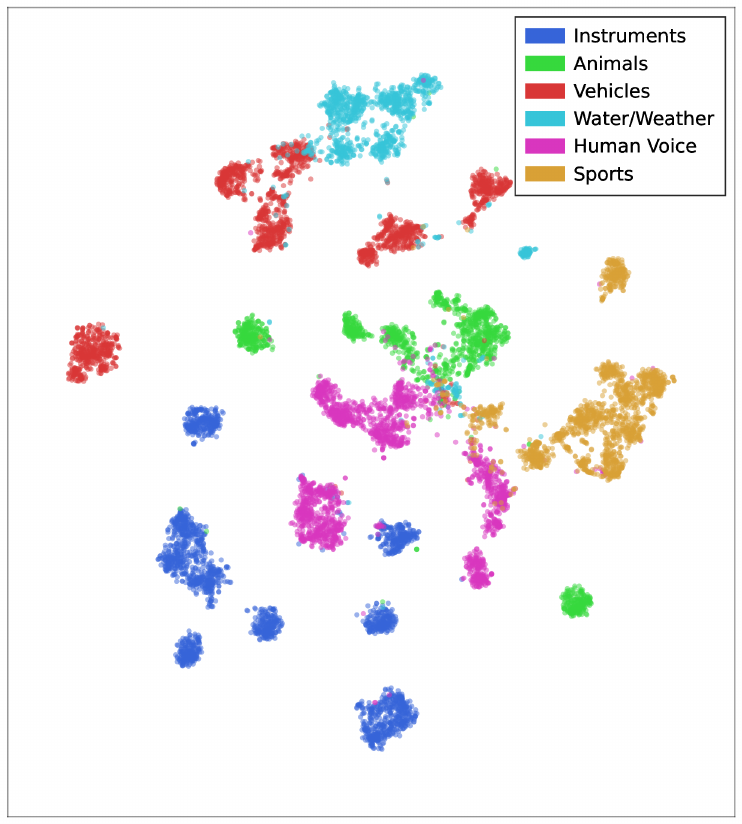}
  \caption{\textbf{Semantic structure of the embedding space.} t-SNE of the
  \texttt{[CLS]} embeddings of the fine-tuned \method ViT-B backbone for
  $N{=}11{,}143$ VGGSound training clips drawn from $60$ classes grouped into six semantic
  families (colours). Clips cluster by family with finer per-class sub-structure
  of the same colour; the residual overlap falls mainly between acoustically
  related families (animal calls vs.\ human voice).}
  \label{fig:tsne_family}
\end{figure}

\cref{fig:tsne_family} asks whether the embedding space is organised by semantic content.
We encode $N{=}11{,}143$ VGGSound training clips spanning $60$ classes, grouped into six
coarse semantic families (instruments, animals, vehicles, water/weather, human
voice, and sports), take the \texttt{[CLS]} embedding of each clip from the
fine-tuned \method ViT-B backbone,
and project the L2-normalised embeddings to two dimensions with t-SNE; points
are coloured by family.
Clips organise into compact, well-separated clusters that respect the family
grouping: each family occupies coherent regions of the projected space, and the
finer per-class structure within a family (for example the individual
instruments) is visible as distinct same-colour sub-clusters.
The residual overlap is concentrated between families that are acoustically and
visually related, most notably animal vocalisations and human voice, which
indicates that the learned space is structured by semantics rather than by
low-level modality cues.

\clearpage

\subsubsection{Feature PCA of Video Patch Tokens}
\label{app:pca}

\begin{figure}[H]
  \centering
  \begin{subfigure}[t]{0.48\linewidth}
    \centering
    \includegraphics[width=\linewidth]{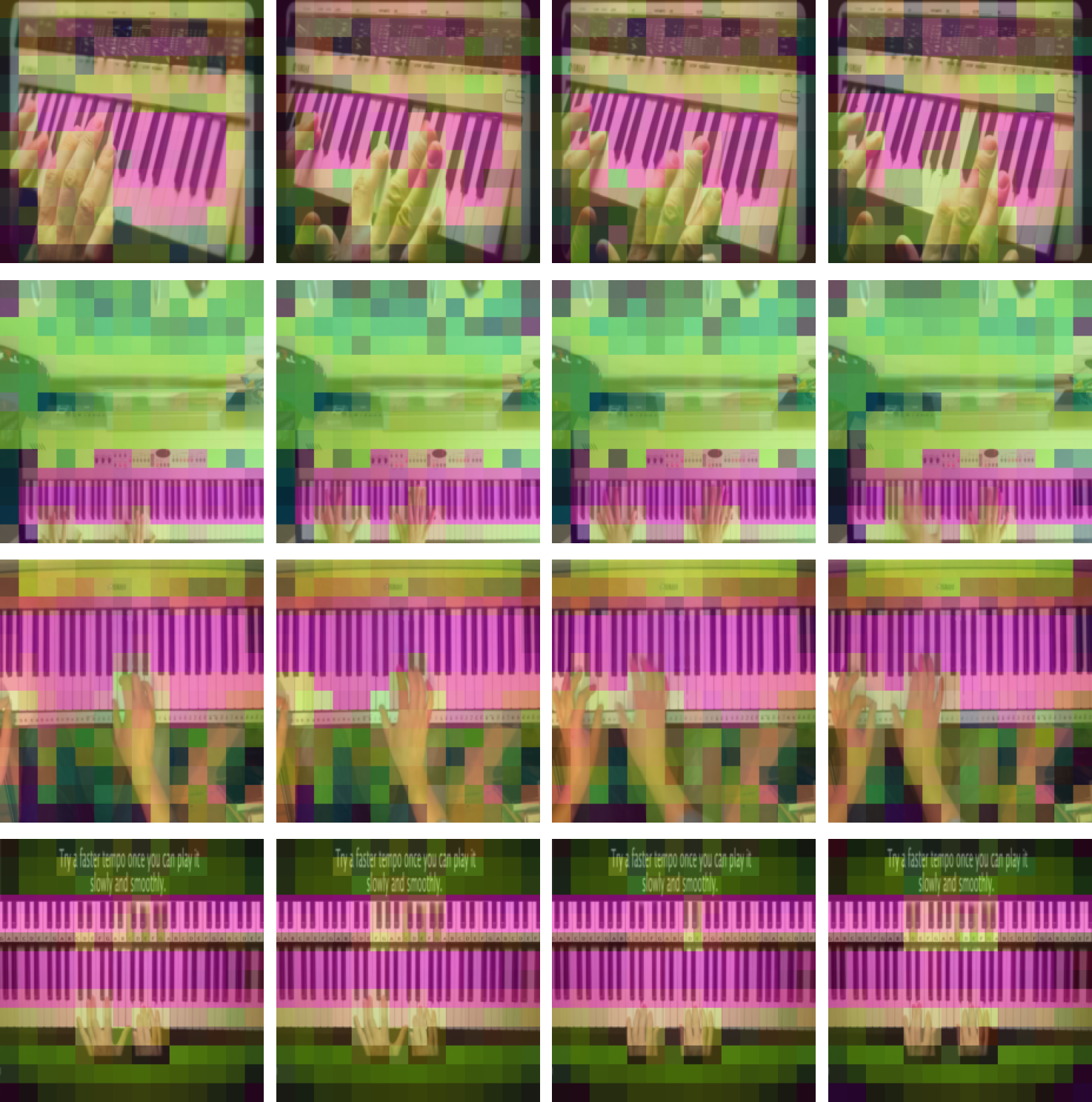}
    \caption{Playing piano}
    \label{fig:pca_piano}
  \end{subfigure}\hfill
  \begin{subfigure}[t]{0.48\linewidth}
    \centering
    \includegraphics[width=\linewidth]{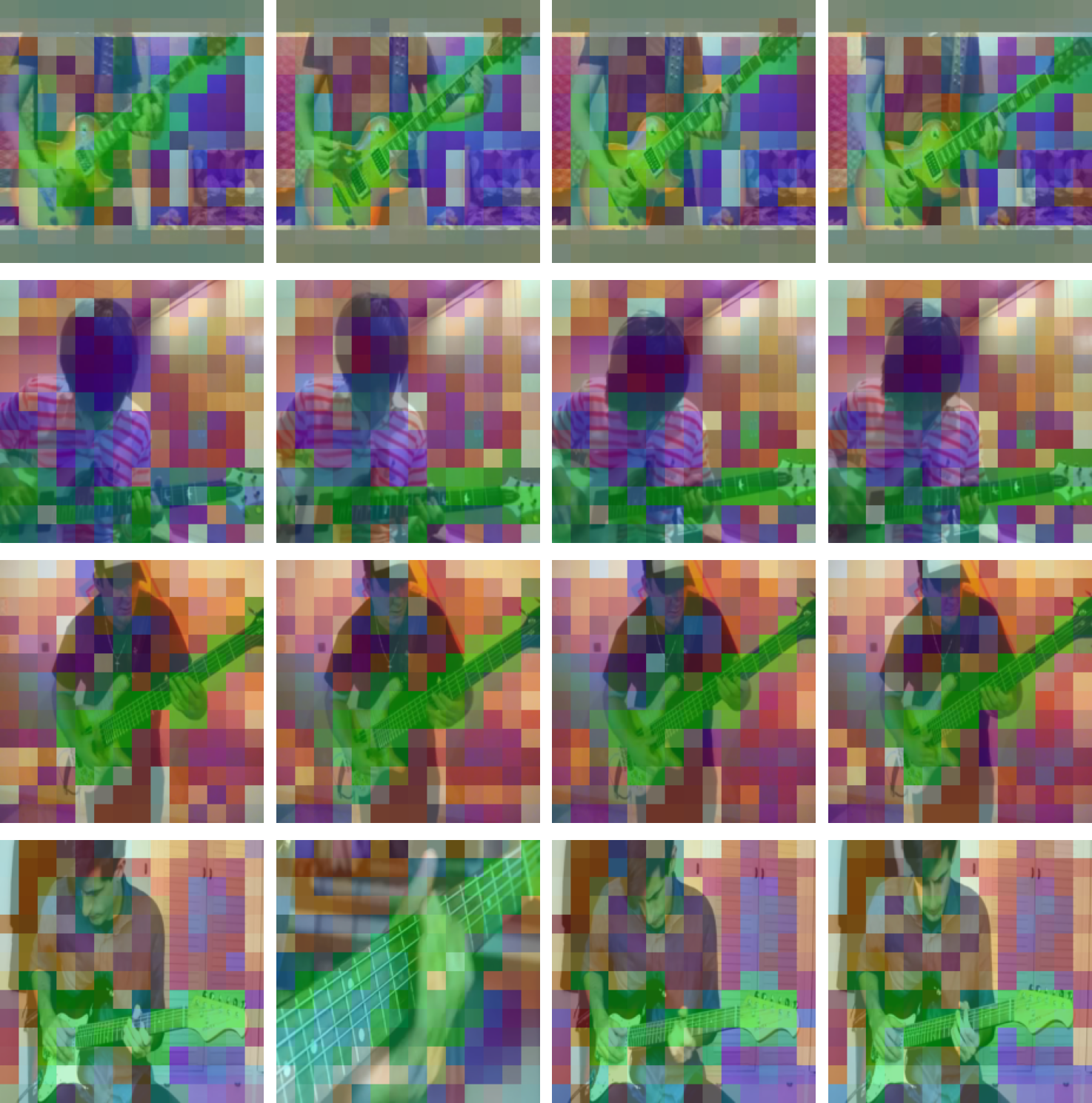}
    \caption{Playing bass guitar}
    \label{fig:pca_bass}
  \end{subfigure}
  \caption{\textbf{Feature PCA of video patch tokens} for two VGGSound
  instrument classes. A single PCA is fit jointly over the last-layer video
  patch tokens of four clips per class (rows); its top three components are
  mapped to RGB and overlaid on four frames per clip (columns). The instrument
  takes a consistent colour across clips and frames, distinct from the player
  and the background.}
  \label{fig:pca}
\end{figure}

\cref{fig:pca} visualises the video patch-token features of the fine-tuned
\method ViT-B encoder in the style of DINO/DINOv2 feature PCA.
For each instrument class we fit a \emph{single} PCA jointly over the pooled
last-layer video patch tokens of four VGGSound clips, map the top three
principal components to RGB, and overlay the result on the video frames (rows
are clips, columns are frames; audio tokens are left uncoloured); because the
basis is shared across the four clips, a given colour denotes the same direction
in feature space. The sounding object (the piano keyboard, the body and neck of
the bass guitar) then takes a consistent colour across clips and
frames, clearly separated from the player and the background, indicating that
the encoder represents the visually salient sound source as a coherent,
instance-corresponding region of feature space, even though it is trained only
with clip-level objectives and never receives pixel- or region-level
supervision, providing a qualitative view of spatial feature structure; the audio-conditioning control is discussed in the main text.

\ifpreprint\else
\clearpage

\subsection{Additional Benchmark Comparisons}
\label{app:finetuned}

This subsection collects the fully fine-tuned results that the main text
summarizes in one sentence each, together with two VGGSound evaluations
that vary the probe and the training budget.

\paragraph{Fully fine-tuned results.}
\cref{tab:finetuned_reported} lists the end-to-end fine-tuned AudioSet-20K
and ESC-50 results of the baselines in \cref{tab:frozen}, as reported in
the respective papers, next to our fine-tuned ViT-B. On AudioSet-20K the
ViT-B reaches 29.6 mAP with joint input, 23.7 with audio only, and 10.3
with video only; joint input outperforms either modality alone, which
establishes complementary visual information without establishing that the
remaining asymmetry is caused by training duration. On ESC-50 the ViT-B
is fine-tuned for 60 epochs per fold with a linear head, head learning
rate $10^{-3}$, backbone multiplier 0.1, SpecAugment with 48 frequency
and 150 time-frame masks, random time shift, and no video tokens (audio-only input);
we report final-epoch accuracy averaged over the official five folds.
Fine-tuned and frozen rows are not equivalent evaluations, and the
fine-tuned baselines remain well ahead in both benchmarks.

\begin{table}[H]
\centering\footnotesize
\setlength{\tabcolsep}{8pt}
\caption{Full-model fine-tuning results. Baseline metrics are as reported in
the respective papers and baseline parameter counts follow
\citet{teotia2026mjepa}; \method rows are our AudioSet-pretrained ViT-B.
AudioSet-20K entries are mAP for audio-only (A), video-only (V), and
audio-video (A-V) inputs; ESC-50 is five-fold accuracy. Best baseline
result per column in bold.}
\label{tab:finetuned_reported}
\begin{tabular}{lllcccc}
\toprule
& & & \multicolumn{3}{c}{AudioSet-20K (mAP)} & ESC-50 \\
\cmidrule(lr){4-6}\cmidrule(lr){7-7}
Method & Params & Pre-train data & A & V & A-V & Acc. \\
\midrule
CAV-MAE~\citep{gong2023cavmae}   & 170M & IN+AS & 37.70 & 19.80 & 42.00 & -- \\
MAViL~\citep{huang2023mavil}     & 170M & IN+AS & 41.80 & 24.80 & 44.90 & 94.4 \\
XKD~\citep{sarkar2024xkd}        & 170M & AS    & --    & --    & --    & \textbf{96.5} \\
EquiAV~\citep{kim2024equiav}     & 170M & IN+AS & \textbf{42.40} & \textbf{25.70} & \textbf{46.60} & 96.0 \\
\method ViT-B (ours)             & 87M  & AS    & 23.7 & 10.3 & 29.6 & 91.4 \\
\bottomrule
\end{tabular}
\end{table}

\subsubsection{Evaluation and Training-Budget Comparisons}

\paragraph{Matching the frozen probe of CAV-MAE Sync.}
We compare a ViT-B backbone pretrained on AudioSet for 25 epochs with
CAV-MAE Sync~\citep{araujo2025cavmaesync} using a probe with two
transformer layers followed by a \texttt{[CLS]} classifier. The backbone
remains frozen while the probe is fitted for 10 VGGSound epochs.
\method reaches 52.96\% top-1 and 79.71\% top-5, compared with the
published 52.7\% top-1. A linear probe on the same backbone reaches
47.43\%. This is comparable performance under matched probe structure and
epoch budgets, rather than evidence of an improvement: our evaluation
averages six temporal crops and uses 8-second audio, whereas the baseline
uses an extended forward pass and 4-second audio. Using a single crop
reduces our result by 0.72 percentage points.

\paragraph{Training duration on VGGSound.}
Extending ViT-B training from 15 pretraining and 3 fine-tuning epochs to
57 and 20, together with a fine-tuning configuration using a backbone
learning-rate multiplier of 0.1 and modality dropout 0.25, increases
attentive-head top-1 from 53.9\% to 60.4\% (58.3\% with the linear
head), the values reported in \cref{tab:main}. Since the training budget
and the fine-tuning configuration both change, the improvement cannot be
attributed to pretraining duration alone.

\paragraph{AudioSet-20K probe fits.}
\label{app:audioset_probe_fits}
The ViT-L audio-video result of 36.03 mAP in \cref{sec:intro} and
\cref{tab:frozen} follows the evaluation protocol of
MJEPA~\citep{teotia2026mjepa}. It uses the ViT-L backbone pretrained for
31 epochs and a frozen depth-4 attentive probe fitted for 40 epochs with
learning rate $3\times10^{-4}$, weight decay $10^{-4}$, batch size 128,
and four-clip logit averaging on 19,446 evaluation videos. A refit
reaches 35.54 mAP. These are distinct fits, not multi-seed confidence
intervals.
\fi

\clearpage

\subsection{Ablations and Hyperparameter Sensitivity}
\label{app:ablations}

We complement the main results with one architectural ablation and two
hyperparameter sensitivity studies, all run as VGGSound-only pretraining
with the same ViT-B early-fusion training configuration as the controlled run reported in
the main text.
The dual-encoder ablation is run on 8$\times$ NVIDIA H200 to match the
controlled main-text run; the SIGReg-weight $\lambda$ and local-view
$K$ sweeps are run on 1$\times$ NVIDIA H200 (batch size 32 per GPU) under a
tighter compute budget.
All curves are clipped to the shortest variant's final
epoch, ensuring a common training budget.

\paragraph{Partial-view ablation (full table).}
\cref{tab:views} lists the local-view variants of \cref{fig:components} with top-5
accuracy, invariance loss, and embedding standard deviation.
All variants use the same ViT-B configuration (effective batch size 512, learning
rate $5\times10^{-4}$, $\lambda{=}0.05$, two global and two local views)
and are compared at epoch 30, with probe accuracies on the held-out test set and
losses averaged over the preceding half epoch.
Both loss-term variants of \cref{fig:components} sit at the floor throughout.

\begin{table}[H]
\centering\small
\caption{View ablation on VGGSound at epoch 30. Probe entries are online top-1/top-5 test accuracy; results are single runs.}
\label{tab:views}
\resizebox{\linewidth}{!}{\begin{tabular}{lrrrr}
\toprule
Pretraining views & Linear & Attentive & Invariance & Embed. std \\
\midrule
Audio-only and video-only local (\method) & 43.9 / 72.7 & 47.1 / 74.9 & 0.042 & 0.989 \\
Random modality dropout ($p{=}0.5$) & 38.1 / 67.7 & 41.3 / 71.2 & 0.060 & 0.966 \\
Both modalities + masking & 23.1 / 49.2 & 34.1 / 63.1 & 0.018 & 0.982 \\
Both modalities in local views & 13.7 / 33.5 & 24.7 / 51.2 & 0.004 & 0.984 \\
No local views ($K{=}0$) & 10.3 / 26.9 & 23.2 / 48.2 & 0.003 & 0.985 \\
Audio-only pretraining & 14.1 / 35.1 & 22.5 / 48.5 & 0.005 & 0.984 \\
\bottomrule
\end{tabular}}
\end{table}

\begin{figure}[H]
\centering
\includegraphics[width=0.9\linewidth]{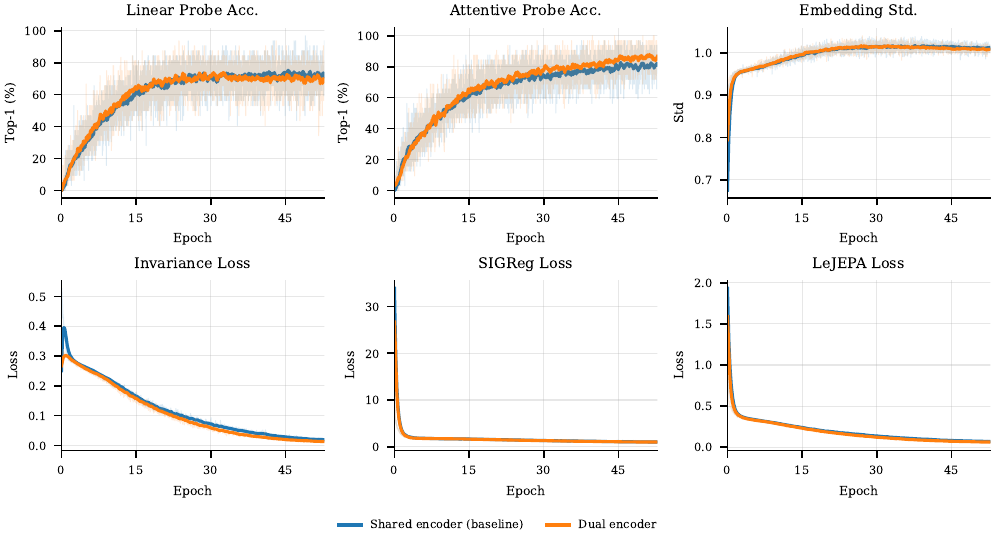}
\caption{
    \textbf{Architectural ablation: shared vs.\ dual encoder} on
    VGGSound pretraining.
    Linear-/attentive-probe top-1 accuracy, embedding standard deviation,
    invariance loss, SIGReg loss, and total LeJEPA loss are plotted
    against pretraining epoch.
    Curves are clipped to the shorter of the two runs (53 epochs).
}
\label{fig:abl_dual_encoder}
\end{figure}

\paragraph{Shared vs.\ dual encoder (ablation).}
\cref{fig:abl_dual_encoder} compares the shared early-fusion encoder used
in \method against a dual-encoder variant that
processes audio and video through separate ViT-B encoders and averages the
two \texttt{[CLS]} tokens.
The shared encoder was comparable to the dual-encoder variant in this single-run comparison on both linear- and
attentive-probe accuracy throughout training; we adopt early fusion for
its simpler, single-encoder design.

\clearpage

\subsubsection{Hyperparameter Sensitivity}

\begin{figure}[H]
\centering
\includegraphics[width=0.72\linewidth]{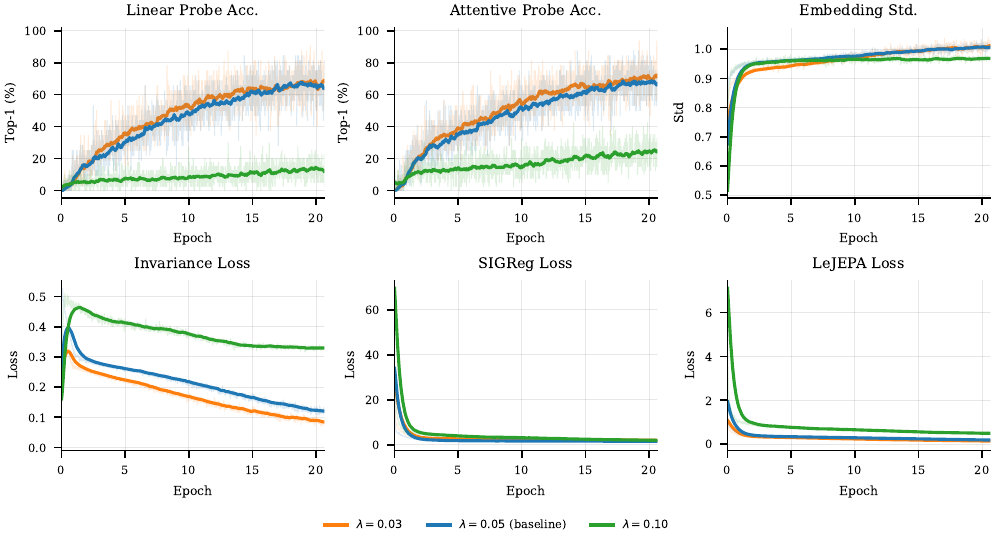}
\caption{
    \textbf{LeJEPA hyperparameter sensitivity: SIGReg weight $\lambda$}
    on VGGSound pretraining, sweeping $\lambda\in\{0.03, 0.05, 0.10\}$
    with $\lambda{=}0.05$ as the baseline used elsewhere in the paper.
    Probe accuracies are from the online probes on the training split.
    $\lambda{=}0.03$ and $0.05$ behave almost identically, whereas
    $\lambda{=}0.10$ leaves both probes far lower and the invariance loss
    far higher.
    Curves are clipped to the shortest run ($\lambda{=}0.10$,
    21 epochs).
}
\label{fig:hp_lambda}
\end{figure}

\paragraph{SIGReg weight $\lambda$ (LeJEPA hyperparameter).}
\cref{fig:hp_lambda} sweeps the SIGReg weight $\lambda$ in the LeJEPA
loss
$\mathcal{L} = (1-\lambda)\mathcal{L}_{\mathrm{inv}} +
\lambda\mathcal{L}_{\mathrm{SIGReg}}$
across $\{0.03, 0.05, 0.10\}$.
$\lambda{=}0.03$ and $\lambda{=}0.05$ produce nearly identical probe-accuracy
curves, both reaching roughly 65--70\% training-split linear and attentive top-1 by epoch 21.
Increasing the weight to $\lambda{=}0.10$ degrades the representation sharply:
the linear probe stays near 13\% and the attentive probe near 24\%, and the
invariance loss plateaus around 0.33 instead of decreasing towards 0.1.
The embedding standard deviation remains close across settings, so the
failure at $\lambda{=}0.10$ is a loss of view alignment rather than collapse.
\method is therefore insensitive to $\lambda$ at or below the default but
sensitive to larger SIGReg weights.

\begin{figure}[H]
\centering
\includegraphics[width=0.72\linewidth]{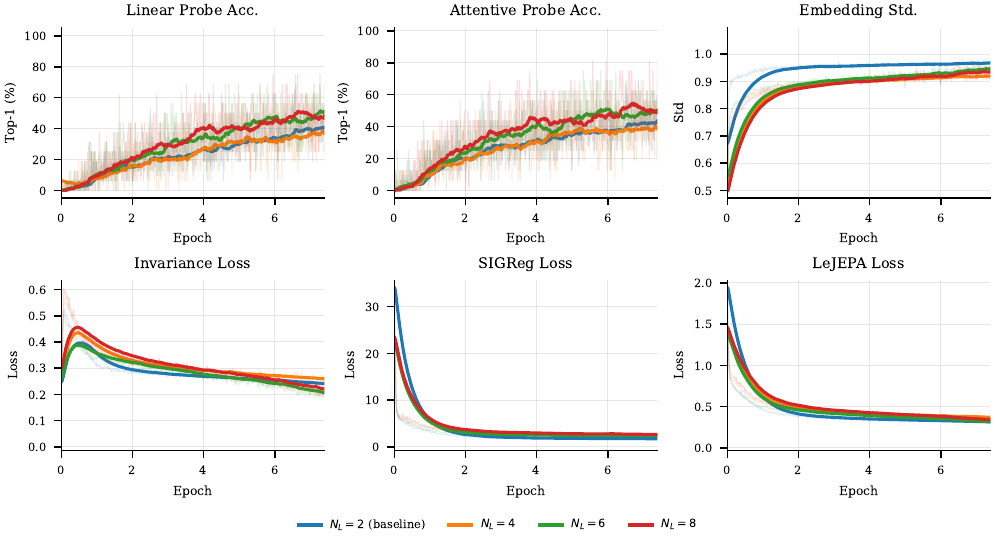}
\caption{
    \textbf{LeJEPA hyperparameter sensitivity: number of local views
    $K$} on VGGSound pretraining, sweeping
    $K\in\{2, 4, 6, 8\}$ with $K{=}2$ as the paper's default.
    Curves are clipped to the shortest run ($K{=}8$, 7 epochs).
}
\label{fig:hp_local_views}
\end{figure}

\paragraph{Number of local views $K$ (LeJEPA hyperparameter).}
\cref{fig:hp_local_views} sweeps the number of local (modality-dropout)
views $K$ across $\{2, 4, 6, 8\}$ while keeping the two global views
fixed.
The linear and attentive probes are largely
insensitive to $K$, suggesting that the default of two local views
already provides sufficient cross-modal regularisation; we did not see a
clear benefit from increasing $K$.

\clearpage

\subsection{Loss-Component and Pipeline Ablations}
\label{app:loss_ablations}

We additionally run three hyperparameter ablations of the VGGSound training procedure, each
with one change from the 1-GPU ablation baseline (ViT-B, batch size 32,
$\lambda{=}0.05$, modality dropout and video tubelet/frequency-time masking,
attentive probe).
All runs share the same wall-clock budget and reach 7 epochs of
VGGSound pretraining; figures are clipped to the common range.
The probe accuracies of the two loss-term ablations are reported in
\cref{fig:components} (right), from longer reruns; here we plot the LeJEPA loss components and the
embedding standard deviation, which expose the mechanism: with
$\lambda{=}0$ the embedding collapses to a point and with $\lambda{=}1$
views never align.

\subsubsection{Removing the SIGReg term (\texorpdfstring{$\lambda{=}0$}{lambda = 0})}
\begin{figure}[H]
\centering
\includegraphics[width=\linewidth]{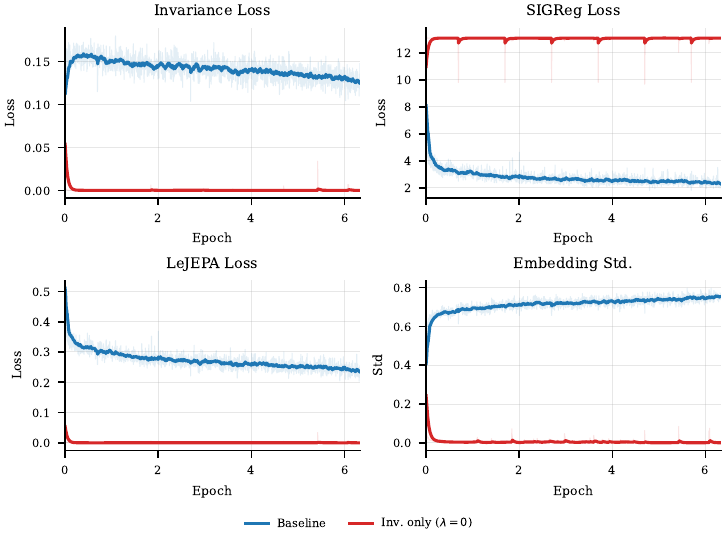}
\caption{
    \textbf{Removing SIGReg ($\lambda{=}0$)} vs.\ the baseline
    ($\lambda{=}0.05$) on VGGSound pretraining.
    Without the SIGReg term, the invariance loss collapses to zero and
    the embedding standard deviation drops several orders of magnitude
    below the baseline, indicating representation collapse.
}
\label{fig:abl_inv_only}
\end{figure}

\cref{fig:abl_inv_only} sets $\lambda{=}0$, training only on the
invariance loss without the SIGReg regulariser.
The invariance loss drops to ${\sim}10^{-7}$ almost immediately and the
embedding standard deviation collapses from the baseline trajectory
towards ${\sim}10^{-4}$, both consistent with the trivial constant-output
solution.
The SIGReg loss itself, although not optimised, rises far above the
baseline range, confirming that the resulting embedding distribution is
arbitrarily far from the target isotropic Gaussian.
This isolates SIGReg as the component that prevents representation
collapse in the audio-visual setting, mirroring the role it plays in
LeJEPA on images.

\clearpage

\subsubsection{Removing the invariance term (\texorpdfstring{$\lambda{=}1$}{lambda = 1})}
\begin{figure}[H]
\centering
\includegraphics[width=\linewidth]{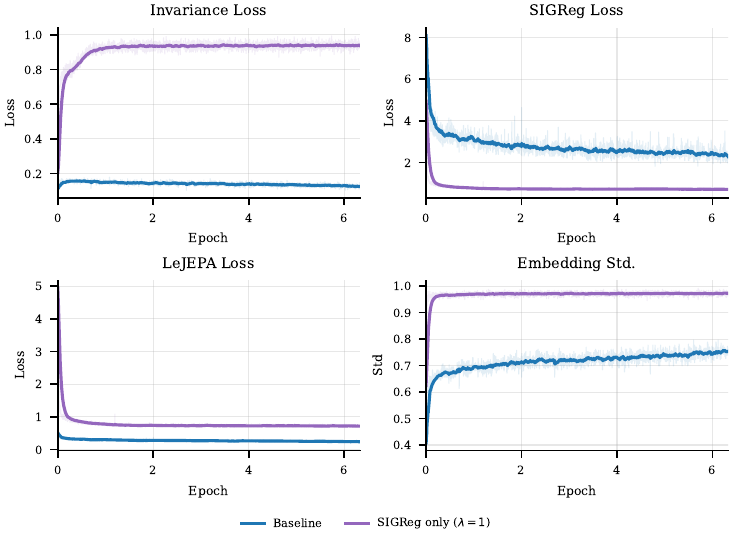}
\caption{
    \textbf{Removing the invariance loss ($\lambda{=}1$)} vs.\ the
    baseline ($\lambda{=}0.05$) on VGGSound pretraining.
    SIGReg alone drives the embedding standard deviation to its target,
    but the invariance loss never decreases, indicating that the model
    never learns to align global and local views without the invariance
    term.
}
\label{fig:abl_sigreg_only}
\end{figure}

\cref{fig:abl_sigreg_only} sets $\lambda{=}1$, training only on SIGReg.
The embedding standard deviation reaches the target ${\sim}1$, confirming
that SIGReg alone can shape the distribution, but the invariance loss
stays roughly an order of magnitude above the baseline throughout
training: there is no signal pulling different views of the same clip
together.
This isolates the invariance loss as the component responsible for
cross-modal alignment.

\clearpage

\subsubsection{Removing modality dropout and masking}
\begin{figure}[H]
\centering
\includegraphics[width=\linewidth]{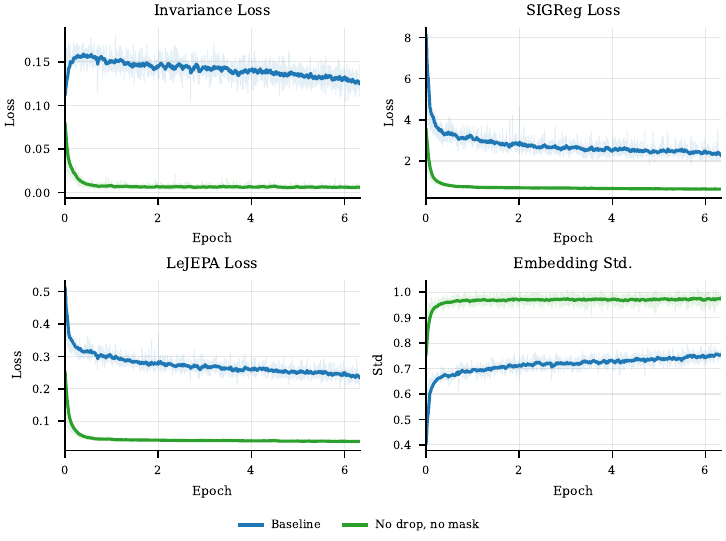}
\caption{
    \textbf{Removing modality dropout and tube/freq-time masking} vs.\
    the baseline on VGGSound pretraining.
    Without partial-view perturbations the invariance loss collapses to
    near zero, indicating that alignment between global and local views
    becomes trivial when both modalities are always present and
    unmasked.
}
\label{fig:abl_no_drop_no_mask}
\end{figure}

\cref{fig:abl_no_drop_no_mask} disables modality dropout, tube masking,
and frequency/time masking ($p_{\mathrm{drop}}{=}0$,
$r_{\mathrm{vid}}{=}0$, $r_{\mathrm{aud}}{=}0$).
Local views then differ from the global views only through random
resized crops and horizontal flips, while always retaining both
modalities.
The invariance loss drops one to two orders of magnitude below the
baseline trajectory because the alignment task is trivialised: there is
no partial-view perturbation forcing the model to predict the missing
modality from the surviving one.
SIGReg behaves similarly to the baseline, so the embedding distribution
is still well shaped, but the LeJEPA loss as a whole loses its
information bottleneck.
This suggests that removing these partial-view perturbations makes alignment easier; the single-factor modality-dropout controls appear in \cref{sec:view_ablation}.

\end{document}